\documentclass[twocolumn,letterpaper]{IEEEAerospaceCLS}  

\usepackage[]{graphicx}    
\usepackage[usenames, dvipsnames]{xcolor}
\usepackage{multicol}
\usepackage[hidelinks]{hyperref}
\usepackage{subcaption}

\usepackage{array}
\usepackage{booktabs}
\usepackage{tabularx}
\usepackage{multirow}
\usepackage{geometry}
\usepackage{algorithmic}
\usepackage{algorithm}
\usepackage{amsfonts}
\usepackage{etoolbox}
\usepackage{siunitx}
\usepackage[acronyms, shortcuts]{glossaries}

\newtoggle{cameraready}
\toggletrue{cameraready} 

\newacronym[longplural={digital elevation models}]{dem}{DEM}{digital elevation models}
\newacronym[longplural={Permanently Shadowed Regions}]{psr}{PSR}{Permanently Shadowed Region}
\newacronym{lroc}{LROC}{Lunar Reconnaissance Orbiter Camera}
\newacronym{spa}{SPA}{South Pole-Aitken}
\newacronym{gitl}{GITL}{``ground-in-the-loop''}

\newacronym[longplural={convolutional neural networks}]{cnn}{CNN}{convolutional neural network}
\newacronym{sgbm}{SGBM}{Semi-Global Block Matching}
\newacronym{clahe}{CLAHE}{Contrast Limited Adaptive Histogram Equalization}
\newacronym{hed}{HED}{Holistically-Nested Edge Detection}

\newacronym{cboe}{CBOE}{coherent backscattering opposition effect}
\newacronym{shoe}{SHOE}{shadow hiding opposition effect}
\newacronym{osl}{OSL}{Open Shading Language}
\newacronym{brdf}{BRDF}{bidirectional reflectance distribution function}

\newcommand{\ignore}[1]{}  

\newcommand{\eg}{{e.g.}}
\newcommand{\ie}{{i.e.}}

\newcommand\mydots{\makebox[1em][c]{.\hfil.\hfil.}}

\begin{document}
\title{ShadowNav: Crater-Based Localization for Nighttime and Permanently Shadowed Region Lunar Navigation}

\author{%
Abhishek Cauligi*\\
abhishek.s.cauligi@jpl.nasa.gov
\and
R. Michael Swan*\\
robert.m.swan@jpl.nasa.gov
\and
Hiro Ono\\
masahiro.ono@jpl.nasa.gov
\and
Shreyansh Daftry\\
shreyansh.daftry@jpl.nasa.gov
\and
John Elliott\\
john.o.elliott@jpl.nasa.gov
\and
Larry Matthies\\
lhm@jpl.nasa.gov
\and
Deegan Atha\\ 
deegan.j.atha@jpl.nasa.gov
\and
\hspace*{20mm} Jet Propulsion Laboratory, California Institute of Technology \hspace*{20mm} \\
Pasadena, CA 91109, USA
\thanks{*Abhishek Cauligi and R. Michael Swan contributed equally to this work.}
\thanks{\footnotesize 978-1-6654-9032-0/23/$\$31.00$ \copyright2023. California Institute of Technology. Government sponsorship acknowledged.
\newline
The research was carried out at the Jet Propulsion Laboratory, California Institute of Technology, under a contract with the National Aeronautics and Space Administration (80NM0018D0004).
}
}

\maketitle

\thispagestyle{plain}
\pagestyle{plain}

\maketitle

\thispagestyle{plain}
\pagestyle{plain}

\begin{abstract}
There has been an increase in interest in missions that drive significantly longer distances per day than what has currently been performed.
For example, Endurance-A proposes driving several kilometers a day in order to reach its target traverse of 2000 km in 4 years.
Additionally, some of these proposed missions, including Endurance-A and rovers for Permanently Shadowed Regions (PSRs) of the moon, require autonomous driving and absolute localization in darkness.
Endurance-A proposes to drive 1200 km of its total traverse at night.
The lack of natural light available during these missions limits what can be used as visual landmarks and the range at which landmarks can be observed.
In order for planetary rovers to traverse long-ranges, onboard absolute localization is critical to the rover’s ability to maintain its planned trajectory and avoid known hazardous regions.
Currently, the localization performed onboard rovers is relative to the rover’s frame of reference and is performed through the integration of wheel and visual odometry and inertial measurements.
To accomplish absolute localization, a “ground-in-the-loop” (GITL) operation is performed wherein a human operator matches local maps or images from onboard with orbital images and maps.
This GITL operation places a limit on the distance that can be driven in a day to a few hundred meters, which is the distance that the rover can maintain acceptable localization error via relative methods.
Previous work has shown that using craters as landmarks is a promising approach for performing absolute localization on the moon during the day.
In this work we present a method of absolute localization that utilizes craters as landmarks and matches detected crater edges on the surface with known craters in orbital maps.
We focus on a localization method based on a perception system which has an external illuminator and a stereo camera.
While other methods based on lidar exist, lidar is not currently planned for deployment on the current proposed nighttime and PSR missions.
In this paper, we evaluate (1) both monocular and stereo based surface crater edge detection techniques, (2) methods of scoring the crater edge matches for optimal localization, and (3) localization performance on simulated Lunar surface imagery at night.
We demonstrate that this technique shows promise for maintaining absolute localization error of less than 10 m required for most planetary rover missions. 
\end{abstract}

\tableofcontents

\begin{figure}[h]
  \centering
  \includegraphics[width=0.48\textwidth,trim={4.5cm 8.5cm 4.cm 6.25cm},clip]{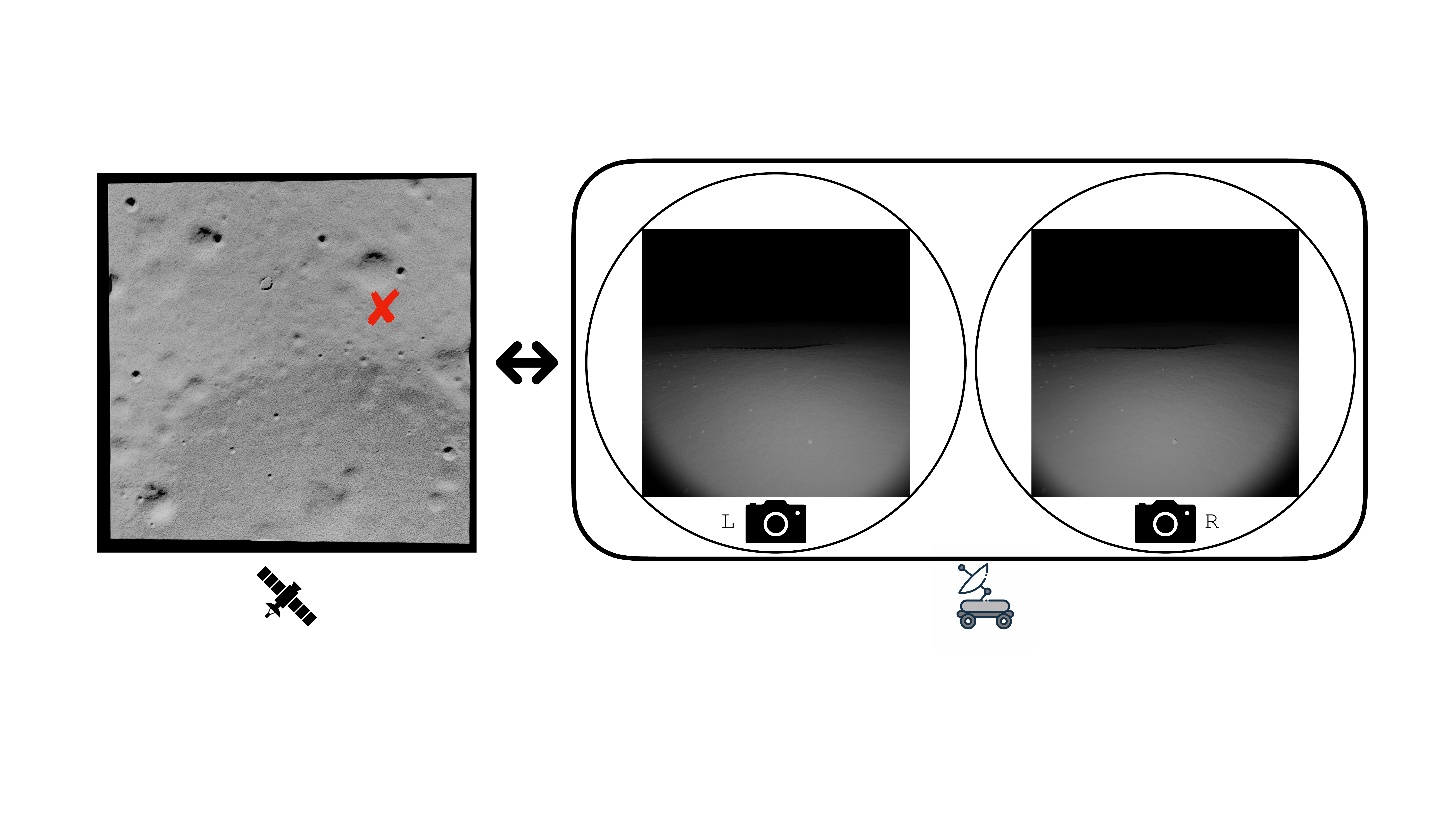}
  \caption{\bf The ShadowNav localization algorithm performs absolute localization for a Lunar rover mission located at the red position in the left image by matching known craters from {\em (left)} an orbital map against {\em (right)} detected craters from the rover stereo cameras.}
  \label{fig:stereo_orbital_overlay}
\end{figure}


\section{Introduction}
\begin{figure*}[t]
  \centering
  \includegraphics[width=\textwidth,trim={4.cm 10cm 4.cm 9.cm},clip]{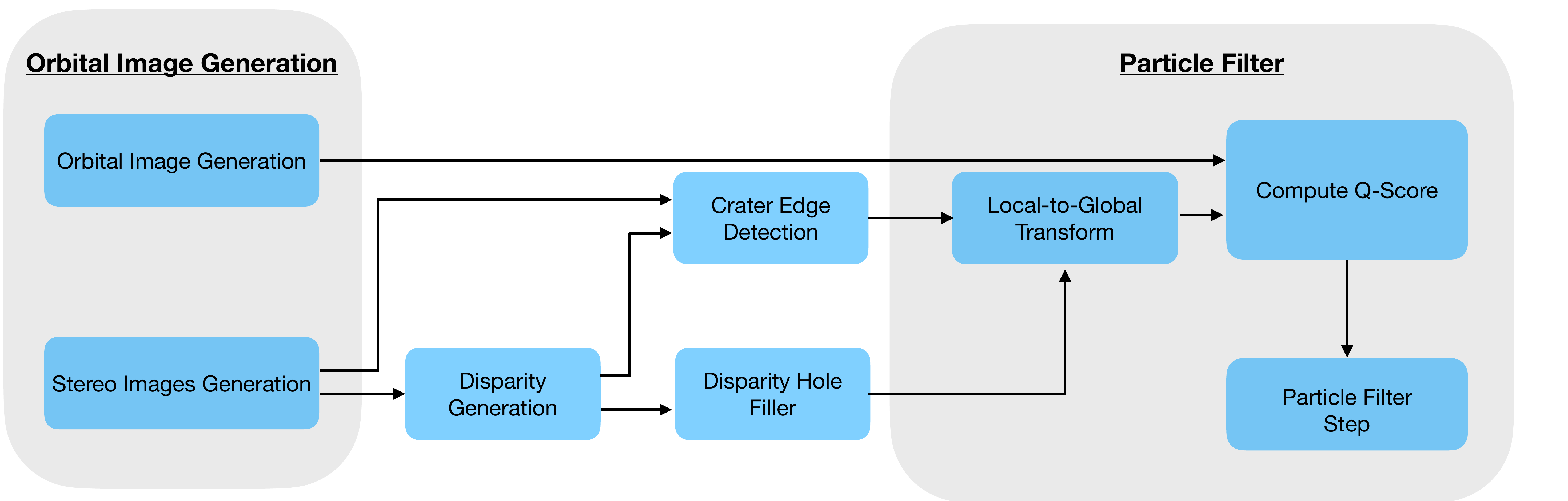}
  \caption{\bf Schematic of the ShadowNav algorithm proposed to perform absolute localization on the Moon.
  A particle filter is used to match craters detected by the rover stero cameras with known craters from an orbital map.}
  \label{fig:shadownav_system_overview}
\end{figure*}

Long-range Lunar navigation, and specifically navigating within darkness, has gained a significant amount of traction recently.
For example, missions to~\acp{psr} of the moon have been proposed such as the VIPER mission~\cite{Ennico-SmithColapreteEtAl2020,ElphicShirleyEtAl2021} and the Lunar Polar Volatiles Explorer mission concepts.
Furthermore, there are missions that have proposed driving during the Lunar night in order to traverse longer distances.
For example, the new Decadal Survey~\cite{NASEM2022} recommends the Endurance-A Lunar rover mission should be implemented as a strategic medium-class mission as the highest priority of the Lunar Discovery and Exploration Program.
The Endurance-A rover proposal plans to drive~\SI{2000}{\kilo\meter} in the~\ac{spa} Basin to collect ~\SI{100}{\kilo\gram} of samples, which would be delivered to Artemis astronauts.
This mission concept study \cite{KeaneTikooEtAl2022} identified several key capabilities required to complete this mission which are: (1) Endurance will need to drive 70\% of its total distance during the night to enable daytime hours dedicated to science and sampling.
(2) The mission will require onboard autonomy for the majority of its operations, while the ground only handles contingencies. (3) Global localization is necessary to maintain an error of $<$\SI{10}{\meter} relative to orbital maps. 

At present, existing rovers perform onboard localization relative to their own reference frame.
This is accomplished by using wheel and visual odometry and inertial measurements. 
Absolute localization is performed periodically with a~\ac{gitl} operation.
This is acceptable for current driving distances which are a few hundred meters a day. 
Existing relative localization has around 2\% drift and therefore can only drive at most \SI{500}{\meter} before the error will be larger than \SI{10}{\meter}.
In order to traverse longer distances, on the order of several kilometers a day proposed by missions such as Endurance-A, autonomous absolute localization becomes critical.
At present the Lunar surface does not have continuous communication with Earth. 
Therefore, having to perform several~\ac{gitl} operations for absolute localization in a day will significantly reduce the distance that can be driven.
The lack of frequent absolute localization for the rover would lead to errors greater than the maximum \SI{10}{\meter} localization error which would present significant risks to the mission through deviations from the desired trajectory and risk for unidentified obstacles.  

Craters as landmarks have been shown to be promising for absolute localization on the Moon~\cite{MatthiesDaftryEtAl2022,DaftryEtAl2023}.
However, the lack of natural light available while driving within a~\ac{psr} or during the Lunar night limits what can be used as a landmark and the range at which the landmarks can be observed.
Using craters is still promising as the average distance between craters of $\geq$\SI{10}{\meter} in diameter is ~\SI{100}{\meter} on terrain with relatively fresh craters and ~\SI{10}{\meter} on terrain with old craters~\cite{HiesingerVanDerBogertEtAl2012}.
Additionally the~\ac{lroc} provides~\acp{dem} with a resolution between \SI{0.5}{\meter}-\SI{5}{\meter} per pixel \cite{RobinsonBrylowEtAl2010} and there are some~\acp{dem} within ~\acp{psr} \cite{CisnerosAwumahEtAl2017}.

In this work, we propose using a stereo camera with an illuminator positioned below the stereo camera in order to detect crater rims within the darkness.
The use of such an illuminator is motivated by the Endurance-A mission concept study~\cite{KeaneTikooEtAl2022}, which proposes the use of a stereo camera with an illumination source as the perception system for a rover operating in darkness.
Global localization is then accomplished by matching the detected crater rims against known craters from an orbital image as shown in Figure~\ref{fig:shadownav_system_overview}.
To handle the uncertainty and nonlinearity of the crater rim detection model, we utilize a particle filter with a novel \emph{Q-Score} metric for ranking potential crater matches in order to estimate the absolute position of the rover within an orbital map.
This paper demonstrates the initial results of both crater detection within darkness and absolute localization within simulation which are the results of the first two years of a planned three year effort to validate this approach.
Work is ongoing to collect and validate this approach in a real-world Lunar analogue test location. 

{\em Statement of Contributions: }This paper presents an approach to absolute localization on the Moon that can be performed while a rover is in darkness, such as within a~\ac{psr} or during the Lunar night. The main contributions of the work as summarized below:

1. We developed a simulator based on Blender~\cite{Blender} which renders simulated surface stereo imagery of the Lunar surface in darkness located within a known orbital position.
The rendering process utilizes the Hapke lighting model for more accurate surface reflectance as well as~\acp{dem} captured by~\ac{lroc} for realistic crater distributions. 

2. We evaluated different crater-edge detection techniques and demonstrate a method which captures 80\% of the leading crater arc at \SI{10}{\meter} and can detect crater arcs out to \SI{20}{\meter}.

3. We present a method to localize a rover within an orbital map using surface crater-edge detections and known orbital craters based on a particle filter and a metric we call the Q-Score which is detailed in Section \ref{subsubsec:q_score}.

4. We demonstrate our absolute localization technique can achieve less than \SI{2}{\meter} absolute error with an assumed odometry drift of 2\% and an initial 3-sigma uncertainty of \SI{3}{\meter}. 

\begin{figure}
  \centering
  \begin{subfigure}{0.23\textwidth}
  \includegraphics[width=\textwidth]{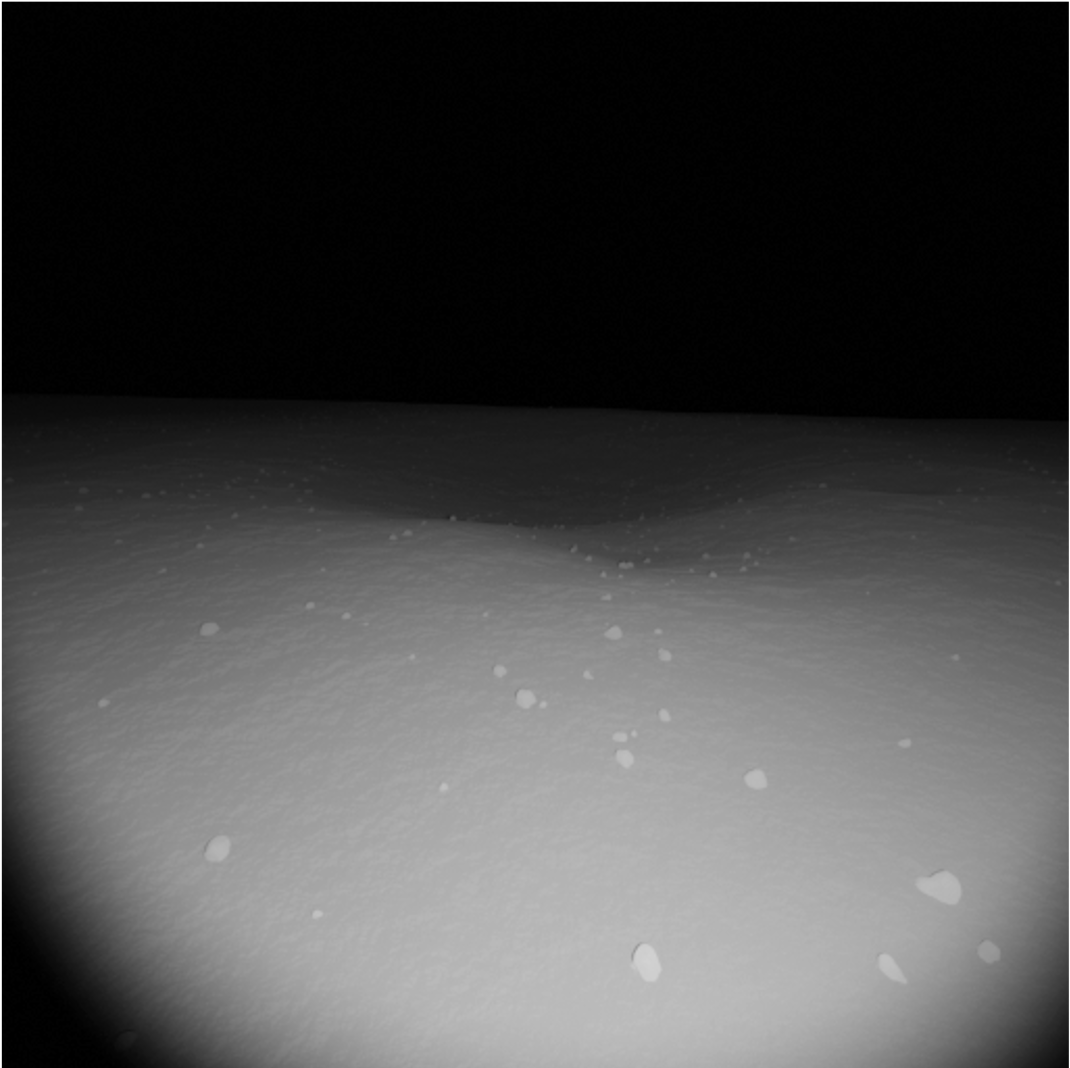}
  \end{subfigure}
  \begin{subfigure}{0.23\textwidth}
  \includegraphics[width=\textwidth]{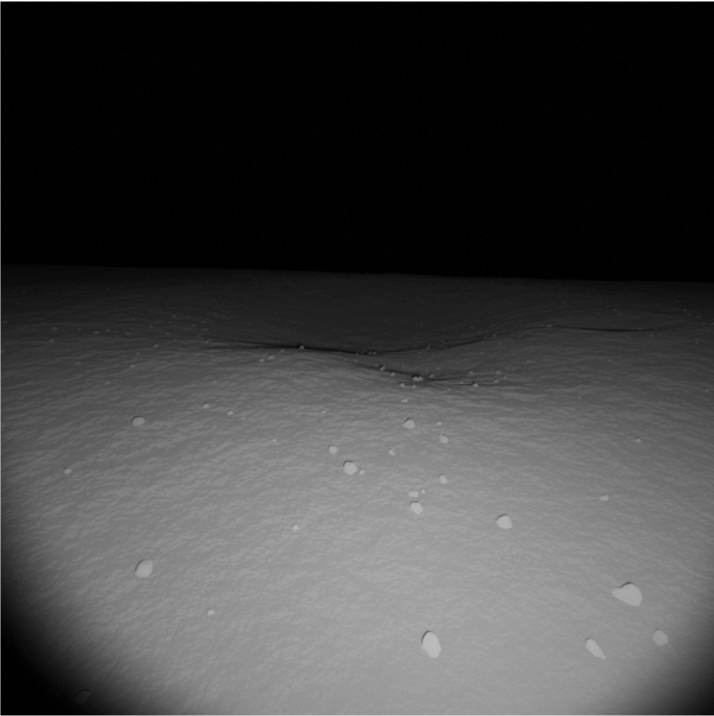}
  \end{subfigure}
  \caption{\bf Figure demonstrating the impact of placement of light source on crater rim shadows. {\em Left:} Sample render of a crater with light source even with camera. {\em Right} Sample render of a crater with light source below the camera.}
  \label{fig:light_source_loc}
  \vspace{-10pt}
\end{figure}

\section{Related Works}

Absolute localization on planetary surfaces is critical for expanding the range rovers can travel in a day and over the course of a mission and there have been many previous works that investigate this problem.
There have been techniques proposed for the Martian surface.
Works such as \cite{hook_mars,ebadi_mars} consider far range and horizon features which are at ranges that are beyond what is expected can be seen in the dark. \cite{HwangboDiEtAl2009} proposes a technique on the Martian surface for absolute localization that uses rocks and~\acp{dem} surface features.

In our work, we focus on the problem of global localization in darkness which is relevant for permanently shadowed regions of the moon, for which there has been a surge of interest in conducting scientific measurements and activities~\cite{HurleyPremEtAl2020}.
Our solution approach is inspired by a host of recent works that seek to leverage orbital maps for global rover localization in these shadowed regions.
In~\cite{HwangboDiEtAl2009}, the authors propose a localization procedure that matches an observed rover image with an orbital map, but this approach neglects the rover motion model and yields a deterministic estimate of the robot belief.
A purely data-driven approach is presented in~\cite{WuPotterEtAl2019}, wherein a convolutional neural network is trained on synthetic data to match the rover observations with orbital imagery.
Closest to our approach,~\cite{FranchiNtagiou2022} presents a particle filtering technique to compare rover monocular camera imagery with orbital imagery and uses a Siamese neural network approach to assign each particle a likelihood weight.
The authors in~\cite{DaftryEtAl2023} propose a similar approach for Lunar absolute localization known as LunarNav.
However, LunarNav focuses on the daytime localization problem and therefore considers different methods of crater matching that rely on greater knowledge of the surface geometry than available in the nighttime case.

\section{Approach}

In this work, we propose an absolute localization approach which utilizes a crater's leading edge as landmarks for localization. 
The end result of this approach will be an estimated position and uncertainty within the orbital frame. 
At present, this approach only considers position localization.
Rover orientation is assumed to be given by a star tracker which can compute orientation in three dimensions from celestial measurements.
Our approach consists of two primary components:
\begin{enumerate}
  \item A leading-edge crater detection methodology for use with a Lunar rover equipped with a stereo camera system and illumination source.
  \item A particle filter for computing a position belief based on a score computed based on the association of crater edges and known orbital ground truth craters, which we call the Q-Score, and the robot motion model.
\end{enumerate}

\subsection{A. Surface Crater Detection}
In order to identify craters on the surface, the system was designed to be used in conjunction with a perception system that contained a stereo camera and an illumination source.
This perception system was configured where the illumination source was beneath the stereo camera.
Examples of simulated images with the light at the same height as the cameras and the light positioned beneath the cameras are in Figure \ref{fig:light_source_loc}.
It was observed that placing the illumination source below the camera results in a shadow at the leading edge of a negative obstacle.
Furthermore, offsetting the light with the cameras reduced the impact of the Hapke model washing out some of the surface texture.
Further details on the Hapke model and its impact on surface terrain are provided in Section~\ref{sec:crater_dataset}.

Here, we first review the three different techniques studied in this work for detecting a crater's leading edge: (1) a method of detecting jumps within stero disparities, (2) a Canny edge detector used to find the shadow on the leading edge, and (3) a~\ac{cnn}-based edge detector that uses both the monocular and disparity image as input.

{\em 1. Stereo Disparity Discontinuity Method}
The first approach for leading edge crater detection relies on detecting discontinuities within the stereo disparity image.
To accomplish this, the stereo disparity image must first be generated using methods such as the JPLV algorithm~\cite{HowardAnsarEtAl2009} or the~\ac{sgbm} approach~\cite{sgbm}, among others.
To account for the low contrast that may be present in the Lunar rover case,~\ac{clahe} is first run on the input images prior to running stereo.
\ac{clahe} is an adaptive histogram equalization and operates on sub-regions of an image which allows more consistent equalization across different lighting conditions within an image. 
This is useful for this application as there is a light-to-dark gradient from near-to-far within the images.
The resulting disparity image is then scanned column-by-column and, when the difference between any two disparities is greater than some pre-defined threshold, the larger column index is marked as a crater edge.
Further, any numerical issues stemming from stereo holes are accounted for by omitting any pixels with spurious values during comparison.

{\em 2. Canny Edge Detector Method}
For sensor configurations that contain an illuminator located beneath the stereo cameras, shadows appear on the leading edge of negative obstacles.
In such cases, a Canny edge detector can be used to distinguish the stark contrasting dark line along the rim.
In this work, the Canny edge detector from OpenCV~\cite{Canny1986} is used to find these shadows. 

{\em 3. \ac{cnn}-Based Edge Detector Method}
The~\ac{hed} approach presents a~\ac{cnn}-based deep learning based method for leading edge crater detection~\cite{XieTu2015}.
This method uses the~\ac{hed} approach and can be performed by directly using the publicly released neural network weights.
\ac{hed} is capable of performing both monocular and stereo depth based edge detection.
For~\ac{hed} to perform edge detection within a depth image, it generates a three channel image that contains horizontal disparity, height above ground, and angle of the local surface normal with the inferred direction of gravity.
The RGB and depth predictions of the~\ac{cnn} are then merged to generate the desired output.

\subsubsection{Positive Obstacle False Positive Rejection}
One shortcoming of the aforementioned leading edge crater detection approach is the susceptibility of false positive cases in the presence of positive obstacles.
In order to account for this positive obstacle issue, the detected edge points are passed through a filter that removes points which have hits on the far side of the crater edge with a detected negative or flat slope.
Detected edge points are kept only if, within the region directly beyond the detected edge, there exists a positive slope or if there is not enough stereo to accurately compute the slope.
Thus, the case of a detected positive slope is assumed to correspond to the rising edge of the crater under the assumption that the detected edge is the leading edge of a negative obstacle.
Alternatively, a detected edge is also retained if the far edge is not captured due to low light conditions, as this is assumed to be an indication of the presence of a large crater.

\begin{algorithm}[t]
\caption{Q-Score Computation}
\label{alg:Qscore}
\algsetup{indent=0.45em} 
\begin{algorithmic}[1]
{\small
    \REQUIRE Belief $b_i^t$, set of crater observations $\{ z_{0,\textrm{rover}}^t, \mydots{}, z_{m,\textrm{rover}}^t \}$, set of ground truth craters $\{ c_{0,\textrm{world}}^t, \mydots{}, c_{\ell,\textrm{world}}^t \}$, positive value $\varepsilon$~\label{line:Qscore_input}
    \STATE $\mathcal{Q}_\textrm{inc} \gets \varepsilon$~\label{line:Qinc_init}
    \FOR {$i=1,\ldots,m$}
        \STATE $z_{0,\textrm{world}}^t \gets \textrm{rover\_to\_world}(z_{0,\textrm{rover}}^t)$~\label{line:tr_meas}
        \STATE $d_\textrm{cr} \gets \min \| c_{j,\textrm{world}}  - z_{0,\textrm{world}}^t\|$~\label{line:cr_association}
        \STATE ${\mathcal{Q}}_\textrm{inc} \gets {\mathcal{Q}}_\textrm{inc} +  d_\textrm{cr} $~\label{line:Qinc_add}
    \ENDFOR
    \STATE $Q_\textrm{score}\gets \min \Big{(} 1, (\frac{1}{m}{\mathcal{Q}}_\textrm{inc})^{-1} \Big{)}$~\label{line:Qscore_comp}
    \RETURN $Q_\textrm{score}$
}
\end{algorithmic}
\end{algorithm}

\begin{algorithm}[t]
\caption{ShadowNav Particle Filtering Algorithm}
\label{alg:ShadowNavPF}
\algsetup{indent=0.45em} 
\begin{algorithmic}[1]
{\small
    \REQUIRE Initial belief distribution $(\mu_0, \Sigma_0)$, number of particles $N_s$, number of effective particles threshold $N_\textrm{eff,thresh}$~\label{line:pf_inputs}
    \STATE $\{ b_1^0, \mydots{}, b_{N_s}^0 \} \gets \textrm{sample\_beliefs}(\mu_0, \Sigma_0)$~\label{line:sample_init}
    \STATE $\{ w_1^0, \mydots{}, w_{N_s}^0 \} \gets \{ 1, \mydots{}, 1 \}$~\label{line:weight_init}
    \STATE $t \gets 1$
    \WHILE{ $\textrm{particle filter running}$}
        \STATE $\{ z_0^t, \mydots{}, z_m^t \} \gets \textrm{get\_observations()}$~\label{line:measurements}
        \STATE $\{ q_1^t, \mydots{}, q_{N_s}^t \} \gets \{ 0, \mydots{}, 0 \}$~\label{line:q_score_init}
        \FOR { $i = 1, \mydots{}, N_s$ }
            \STATE $b_i^{t} \gets \textrm{propagate\_sample}(b_i^{t-1})$~\label{line:belief_update}
            \STATE $q_i^{t} \gets \log\textrm{Q\_score}(b_i^{t}, \{ z_0^t, \mydots{}, z_m^t \})$~\label{line:q_score_comp}
        \ENDFOR
        \STATE $q_\textrm{min}^t \gets \min ( q_1^t, \mydots{}, q_{N_s}^t )$~\label{line:q_score_min}
        \FOR { $i = 1, \mydots{}, N_s$ }
            \STATE $w_i^{t} \gets w_i^{t-1} + q_i^t - q_\textrm{min}^t$~\label{line:weight_update}
        \ENDFOR
        \STATE $N_\textrm{eff} \gets \textrm{compute\_N}_\textrm{eff}( w_1^t, \mydots{}, w_{N_s}^t )$~\label{line:compute_Neff}
        \IF {$N_\textrm{eff} \leq N_\textrm{eff,thresh}$}
            \STATE $\{ b_1^t, \mydots{}, b_{N_s}^t \} \gets \textrm{resample\_beliefs}(\{ b_i^t \}_{i=1}^{N_s}, \{ w_i^t \}_{i=1}^{N_s} )$~\label{line:resample}
            \STATE $\{ w_1^t, \mydots{}, w_{N_s}^t \} \gets \{ 1, \mydots{}, 1 \}$~\label{line:weights_reassign}
        \ENDIF
        \STATE $t \gets t+1$
    \ENDWHILE
}
\end{algorithmic}
\end{algorithm}

\begin{algorithm}[t]
\caption{Systematic Resampling}
\label{alg:systematic_resampling}
\algsetup{indent=0.45em} 
\begin{algorithmic}[1]
{\small
    \REQUIRE Particles $\{ b_1^t, \mydots{}, b_{N_s}^t \}$ and associated weights $\{ w_1^t, \mydots{}, w_{N_s}^t \}$
    \STATE $n^t = \log \Big{(} \sum_{i=1}^{N_s} \exp(b_i^t) \Big{)}$~\label{line:weight_normalization_val}
    \STATE $\{ \tilde{w}_0^t, \mydots{}, \tilde{w}_{N_s}^t \} \gets \{0, \mydots{}, 0 \}$~\label{line:normalized_weights_init}
    \FOR {$i = 1, \mydots{}, N_s$}
        \STATE $\tilde{w}_i^t \gets \exp(w_i^t - n^t)$~\label{line:normalize_weights}
    \ENDFOR
    \STATE $\{ q_0, \mydots{}, q_{N_s} \} \gets \textrm{cum\_sum}(\{ \tilde{w}_0^t, \mydots{}, \tilde{w}_{N_s}^t \})$~\label{line:weights_cumsum}
    \STATE $n \gets 0$
    \STATE $m \gets 0$
    \STATE $u_0 \sim {\mathbb{U}} ( 0, \frac{1}{N_s})$~\label{line:sample_uniform_val}
    \WHILE {$n \leq N_s$}
        \STATE $u = u_0 + \frac{n}{N_s}$
        \WHILE {$q_m \leq u$}
            \STATE $m \gets m+1$
        \ENDWHILE
        \STATE $n \gets n+1$
        \STATE $b_n^t \gets b_m^t$
    \ENDWHILE
    \RETURN $\{ b_0^t, \mydots{}, b_{N_s}^t \}$
}
\end{algorithmic}
\end{algorithm}

\subsection{B. Particle Filter}
Here, we provide an overview of the proposed ShadowNav particle filtering approach.
First, we provide further details on the Q-Score metric that is used in the belief update step.

\subsubsection{Q-Score}\label{subsubsec:q_score}
The Q-Score provided the measurement probability of some position belief based on rover frame observations and an orbital map.
The procedure for computing the Q-Score is given in Algorithm~\ref{alg:Qscore}.
The algorithm takes as input a given belief $b_i^t$, a set of $m$ observed edges in rover frame, and a set of $\ell$ ground truth crater observations to associate these measurements with (Line~\ref{line:Qscore_input}).
A value $\mathcal{Q}_\textrm{inc}$ is initialized to some negligibly small, positive value $\varepsilon$ to later avoid divide-by-zero issues (Line~\ref{line:Qinc_init}).
Next, for each measurement $z_i^t$ in the rover frame, the detected edge is converted to world frame (Line~\ref{line:tr_meas}) and the minimum distance to an edge from the ground truth map computed (Line~\ref{line:cr_association}).
The $\mathcal{Q}_\textrm{inc}$ is incremented by the distance between the observed edge and its associated ground truth observation (Line~\ref{line:Qinc_add}).
The Q-Score is computed as the reciprocal of $\mathcal{Q}_\textrm{inc}$ and a $\min$ operation is applied to ensure that the score provided by any particular run is between 0 and 1 (Line~\ref{line:Qscore_comp}).
This implies that observations and belief pairs which are less than \SI{1}{\meter} away from ground truth will receive the same score as those exactly 1m away from ground truth, which is seen as acceptable given the orbital~\ac{dem} resolution and mission concept localization requirements.

In addition to the shortest distance formulation from Line~\ref{line:cr_association}, additional approaches were also explored for determining the Q-Score.
One alternate approach investigated included fitting a Gaussian normal distribution on the orbital map crater edges and the Q-Score value was them computed based on the intensity (\ie{}, distance to the computed mean) of the point hit by observations or 0 in cases when no point was hit.
In practice, it was determined that the shortest distance formulation provided the most robust results for use with the particle filter and also did not require additional projection calculations to project each belief from the orbital frame to rover frame.

\subsubsection{Overview}
A description of the ShadowNav particle filtering algorithm is given in Alg.~\ref{alg:ShadowNavPF}.
The algorithm takes as input a Gaussian belief distribution $(\mu_0, \Sigma_0)$ assumed for the initial robot position, the number of particles $N_s$ to use in the particle filter, and a threshold for the effective number of beliefs $N_\textrm{eff,thresh}$ used to trigger resampling (Line~\ref{line:pf_inputs}).
The filter is initialized by sampling $N_s$ particles from the initial belief distribution and assigning a weight of equal importance for each particle (Lines~\ref{line:sample_init}-\ref{line:weight_init}).
As common in particle filtering implementations~\cite{ArulampalamMaskellEtAl2002}, we note that we used the $\log$ of the weights for improved numerical stability of the weight update step~\cite{gentner2018log}.
Given a new set of crater observations (Line~\ref{line:measurements}), a set of Q-Score measurements is initialized for computing for each individual particle (Line~\ref{line:q_score_init}).
After applying the motion model update to each particle (Line~\ref{line:belief_update}), the Q-Score for each updated particle is computed using the procedure from Alg.~\ref{alg:Qscore} by comparing against the current measurements (Line~\ref{line:q_score_comp}).
The particle weights are then updated in $\log$-domain (Line~\ref{line:weight_update}) with a normalization step to ensure non-negative weights (Line~\ref{line:q_score_min}).
Next, the number of effective samples $N_\textrm{eff}$ at the current iteration is calculated (Line~\ref{line:compute_Neff}).
A common pitfall of particle filters is ``degeneracy'', wherein the weights $\{ w_i^t \}$ collapse around a handful of particles and computational resources are wasted on propagating low likelihood particles~\cite{ArulampalamMaskellEtAl2002}.
If $N_\textrm{eff}$ is below the threshold $N_\textrm{eff,thresh}$, then this indicates that the filter is degenerating and a resample operation is triggered (Line~\ref{line:resample}).

Further details on the systematic resampling approach used in this work are provided in Algorithm~\ref{alg:systematic_resampling}.
Given a set of particles and their associated weights, the weights are first normalized to $(0, 1]$ from $\log$-domain (Lines~\ref{line:weight_normalization_val}-\ref{line:normalize_weights}) and the cumulative sum of these normalized weights $\tilde{w}_i^t$ computed (Line~\ref{line:weights_cumsum}).
The key step in systematic resampling is to sample a random value $u_0$ from a uniform distribution inversely proportional to $N_s$ (Line~\ref{line:sample_uniform_val}) and then incrementally sample a new particle from this ``bin'' of width $\frac{1}{N_s}$ .
This ensures that, after resampling, at least one particle is retained from each $\frac{1}{N_s}$ interval from the previous belief distribution.

\begin{figure}[t]
  \centering
  \begin{subfigure}[t]{0.15\textwidth}
    \includegraphics[width=\textwidth]{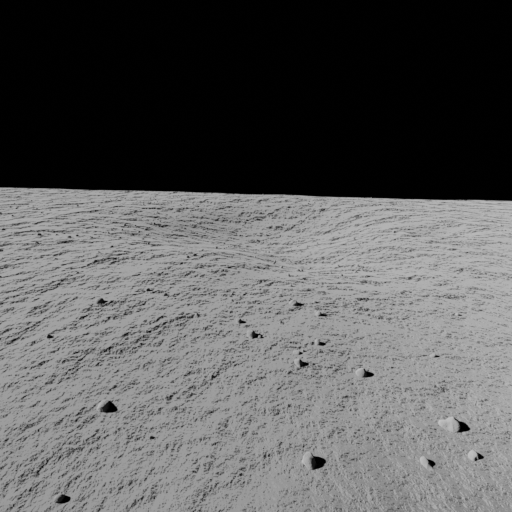}
    \caption{Sample of terrain with 90$^\circ$ from camera.}
  \end{subfigure}
  \begin{subfigure}[t]{0.15\textwidth}
    \includegraphics[width=\textwidth]{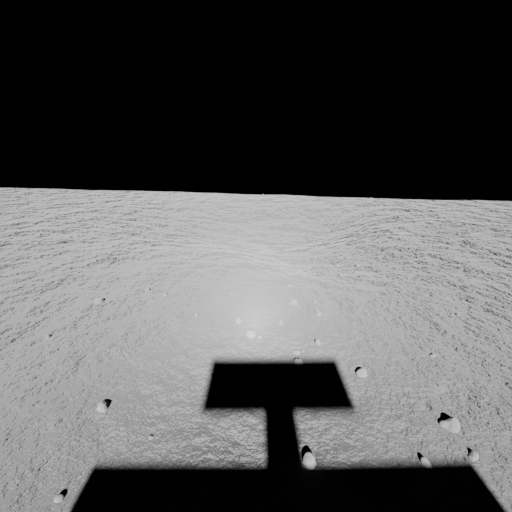}
    \caption{Sample opposition effect during the day.}
  \end{subfigure}
  \begin{subfigure}[t]{0.15\textwidth}
    \includegraphics[width=\textwidth]{fig/sim_details/light_below_cam_left.png}
    \caption{Sample effect of surface reflectance at night with an illuminator.}
  \end{subfigure}
  \caption{\bf The opposition effect simulated during the day and its effect at night with an external illuminator.}
  \label{fig:hapke_effect}
\end{figure}

\subsection{C. Surface to Orbital Crater Transformation}
For every observation step, rover frame crater edges were detected with a stereo camera pair that provided the depth, and thus a relative position for the crater edge was saved.
This relative crater distance was added to each particle's belief position to form an estimate of the observed crater position in the world frame for each particle.
The orbital map was projected to the world frame and then the shortest distance metric noted in the Q-Score algorithm was used to determine which particle belief positions were most likely and thus which observed crater was the most likely one to match the known orbital craters.

\subsubsection{Stereo hole filling}
As some crater edge detections do not rely on depth information, not all pixels in the stero camera depth or disparity image will have a detected depth value and, in such cases, no relative position would be available for matching rover observations to the orbital map.
For such observations, a simple plane fit can be carried out to fill in the depth information.
A future area of investigation includes carrying out an improved stereo hole filling approach, in particular using existing knowledge on what the regional terrain looks like.

\begin{figure}
  \centering
  \includegraphics[width=0.4\textwidth]{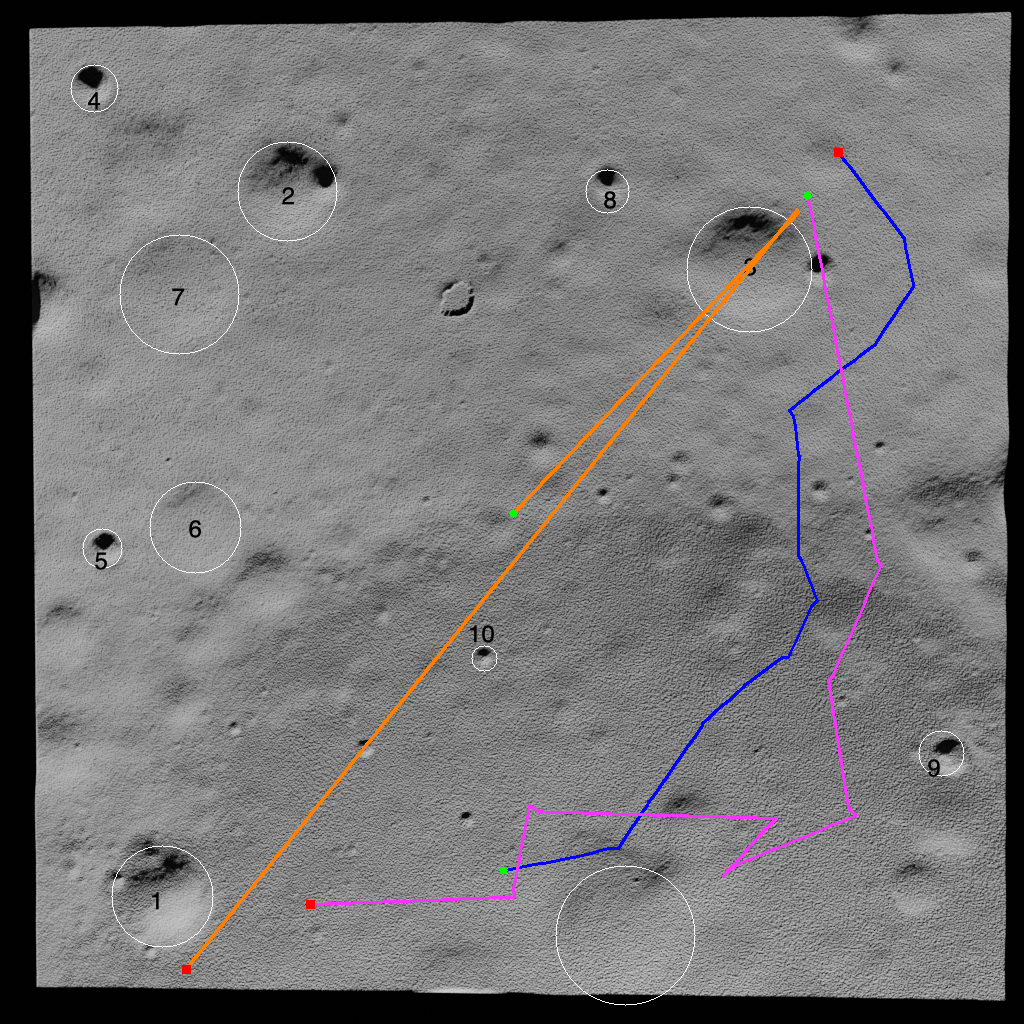}
  \caption{\bf The trajectories used for the numerical experiments are overlaid on the orbital map here with the crater numbers in black.
  A red square indicator is at the start and a green circle indicator is at the end of each trajectory. 
  Trajectory 1 is in blue, trajectory 2 is in orange, and trajectory 3 is in purple.}
  \label{fig:trajectories}
  \vspace{-5pt}
\end{figure}

\section{Datasets Overview}

\subsection{A. Simulated Lunar Environment}
At the time of writing, a Lunar dataset with images captured in the dark with an illuminator did not exist.
Therefore, to evaluate the approach, a simulation environment was developed using the Blender software~\cite{Blender}.
In order to simulate images as realistically as possible, the Hapke lighting model \cite{hapke2012theory,hapke2002bidirectional,schmidt2019efficiency} was implemented.
This model approximates the Lunar surface reflectance and will simulate the ``opposition effect''.
This effect leads to a focused point of extreme saturation at a location within an image where the camera ray and light source are at zero phase angle.
The Hapke lighting model was implemented using the ``old highland'' parameters of the moon provided in \cite{xu2020photometric}, as these most closely match the poles of the moon where~\acp{psr} can be found.
The~\ac{cboe} was left out of our implementation and only the~\ac{shoe} was implemented as it dominates most or all lighting calculations in our use case, while~\ac{cboe} has a negligible or very small effect.
Initial implementation was done using the~\ac{osl}, however not all rays are available for calculation due to optimizations made in~\ac{osl}, so workarounds were needed to implement the Hapke lighting model in Blender using~\ac{osl}.
While this was partially successful, it was not very robust and we had numerous issues.
Instead of using~\ac{osl}, we opted to modify the source of Blender to add the Hapke~\ac{brdf} directly into the Blender Cycles renderer code which also reduced the render time by greater than a factor of 2 through the use of Nvidia CUDA.

In order to represent a realistic 3D model of the surface geometry,~\acp{dem} produced from~\ac{lroc} were utilized.
While~\ac{lroc} has enough resolution to resolve craters of around \SI{10}{\meter}, its resolution is not quite good enough for generating smooth surface imagery.
In order to have smooth surface image renders, the~\acp{dem} from~\ac{lroc} were scaled down to be \SI{0.25}{\meter} resolution.
Crater measurements in future discussions were based on this scaled resolution.
This scaled~\ac{dem} was imported into Blender and a surface texture was added.
The surface texture comprised of two scales of fractal Brownian motion, which is a natural noise that was added to the~\ac{dem} in order to simulate Lunar surface texture for stereo to utilize.
Figure~\ref{fig:hapke_effect} demonstrates three sample renders, two in the daylight and one at night with an illumination source from our simulation.
It demonstrates what the surface looks like in daytime conditions as well as the effect of the Hapke model during the day with the sun behind the camera and the effect of the illumination source.
From this it was observed that the full amount of daytime texture is not observed during the night with an illumination source. 

\begin{table}
\renewcommand{\arraystretch}{1.3}
\caption{\bf Table of crater sizes in crater detection dataset.}
\label{tab:crater_sizes}
\centering
\begin{tabular}{|l|c|c|}
\hline
\bfseries Crater & \bfseries Diameter (m) & \bfseries Depth (m) \\
\hline
\hline
1 & 9.2 & 1.0 \\
2 & 9.1 & 0.75 \\
3 & 11.3 & 0.84 \\
4 & 4.4 & 0.55 \\
5 & 3.7 & 0.40 \\
6 & 8.3 & 0.27 \\
7 & 11.9 & 0.44 \\
8 & 3.9 & 0.48 \\
9 & 4.1 & 0.49 \\
10 & 2.3 & 0.25 \\
\hline
\end{tabular}
\vspace{-10pt}
\end{table}

\subsection{B. Simulated Craters for Detection Analysis}
\label{sec:crater_dataset}

In order to evaluate the performance of different crater detection techniques, a dataset with different sized craters was built.
This dataset was built using the simulation process within Blender  and captured stereo pair renders between \SI{5}{\meter} and \SI{20}{\meter} from the front crater rim in increments of \SI{0.1}{\meter}.
This dataset contained 10 different craters with varying sizes and depths.
The sizes of the craters within this dataset are in Table \ref{tab:crater_sizes} and their locations corresponding to the crater ID in our simulated environment are marked in Figure~\ref{fig:trajectories}. 

\subsection{C. Simulated Trajectories for Localization Analysis}

In order to evaluate the localization performance, several trajectories were run in the simulated environment.
These trajectories were run to generate an image every \SI{1}{\meter} and were designed to approach craters in different ways that might present challenges to our filtering approach.
The \SI{1}{\meter} observation delta was used to reduce render times of our dataset, as rendering every \SI{0.1}{\meter} did not result in a significant localization performance change.
An overview of the trajectories within the orbital environment are displayed in Figure~\ref{fig:trajectories}. 

\subsection{D. Real Data of Negative Obstacles at Night}

In addition to the simulated data generated, a dataset was collected in the Arroyo, which is a dry river bed near the NASA Jet Propulsion Laboratory.
This dataset contained a few different negative obstacles that were imaged at 5, 10, and \SI{15}{\meter} away from the leading edge. 
This dataset was used to validated that the stereo and crater edge detection algorithms work on real data collected at night with an external illuminator.

\begin{figure}
  \centering
  \begin{subfigure}{0.23\textwidth}
  \includegraphics[width=\textwidth]{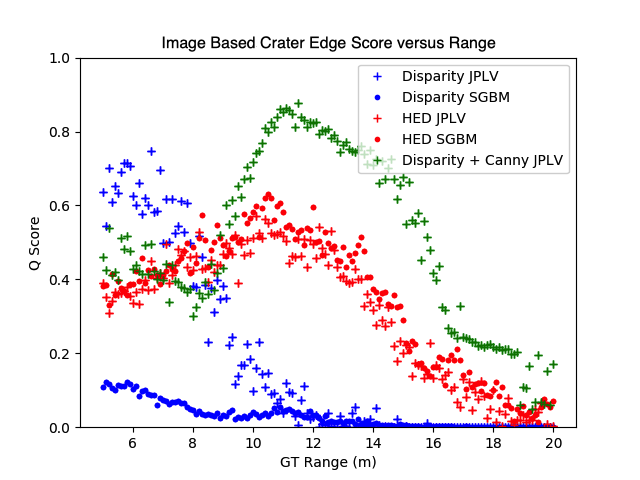}
  \end{subfigure}
  \begin{subfigure}{0.23\textwidth}
  \includegraphics[width=\textwidth]{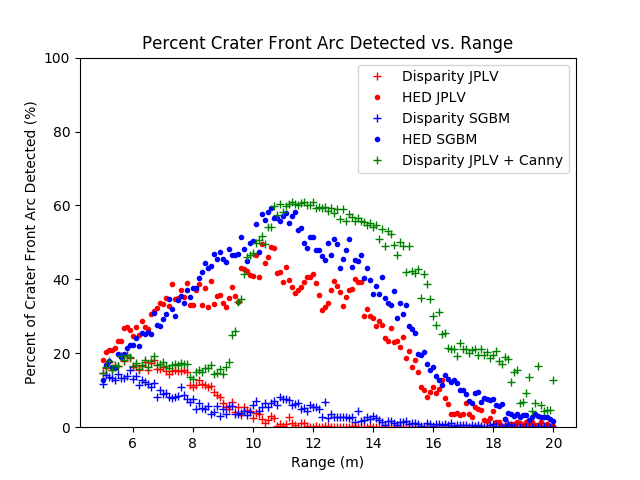}
  \end{subfigure}
  \caption{\bf Plots of different metrics evaluating crater detection performance. {\em Left:} Plot that shows image-based crater edge detection score versus range for all craters evaluated. {\em Right:} Plot that shows percent of the crater front arc detected for all craters evaluated.}
  \label{fig:crater_det}
\end{figure}

\section{Crater Detection Performance}

\begin{figure}
  \centering
  \includegraphics[width=0.46\textwidth]{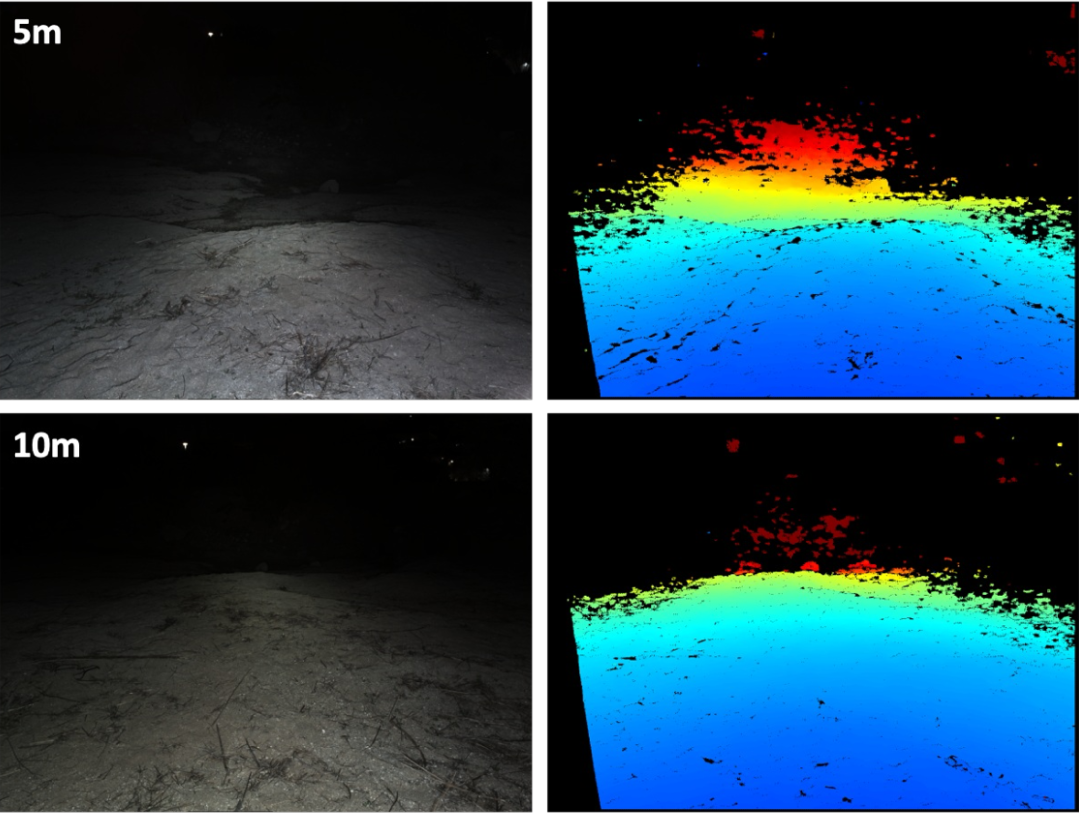}
  \caption{\bf Sample stereo results using JPLV stereo on a sample negative obstacle.}
  \label{fig:arroyo_stereo}
  \vspace{-5pt}
\end{figure}

\begin{figure*}[t!]
    \centering
    \begin{subfigure}[t]{0.32\textwidth}
        \includegraphics[width=\textwidth]{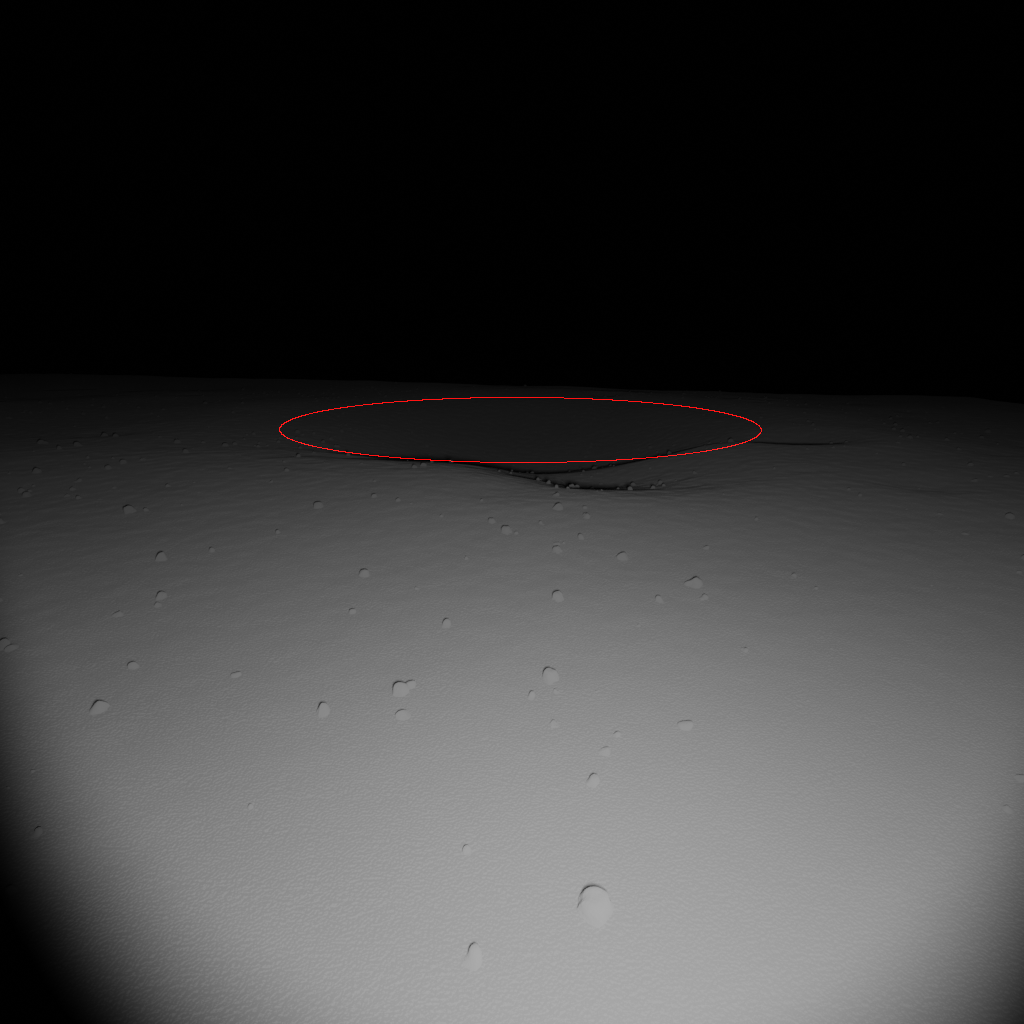}
        \caption{Ground Truth at \SI{7}{\meter}}
        \label{fig:qual_dets_a}
    \end{subfigure}
    \begin{subfigure}[t]{0.32\textwidth}
        \centering
        \includegraphics[width=\textwidth]{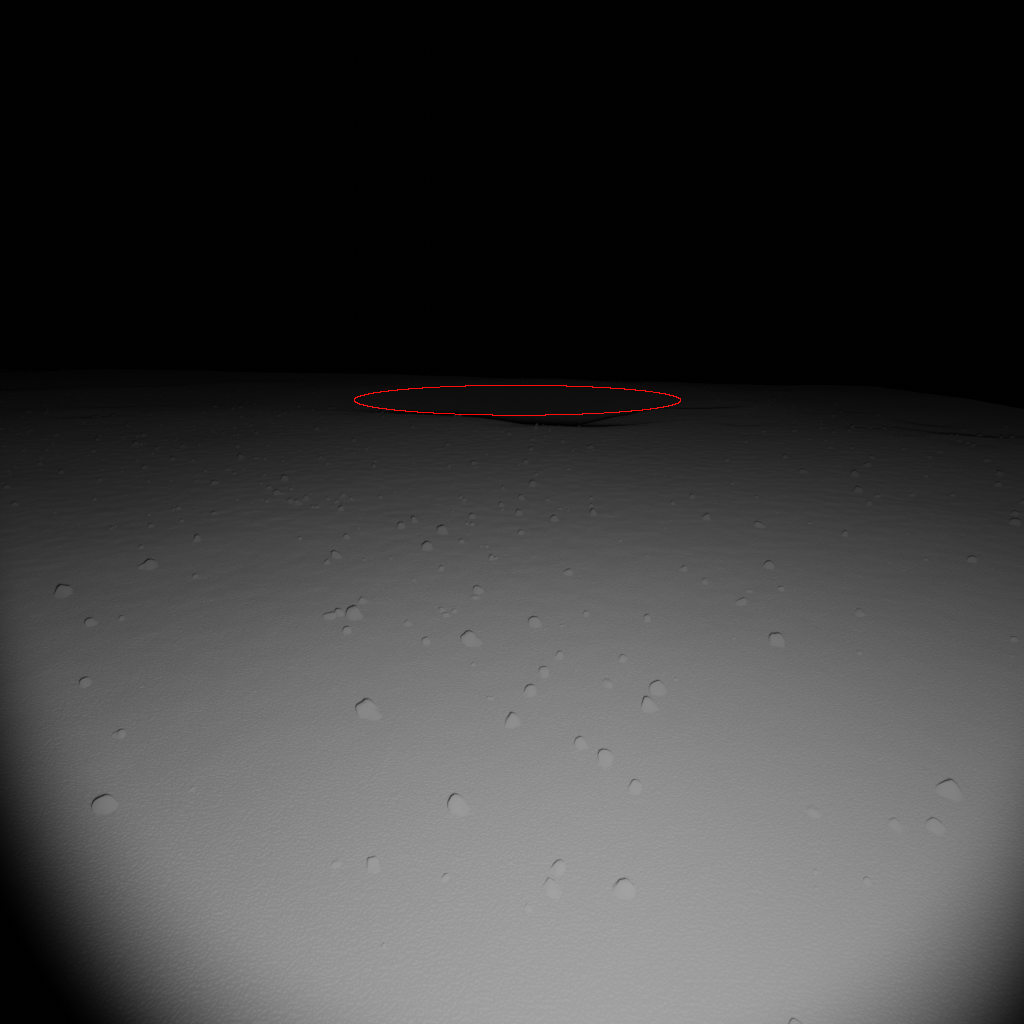}
        \caption{Ground Truth at \SI{12}{\meter}}
        \label{fig:qual_dets_b}
    \end{subfigure}
    \begin{subfigure}[t]{0.32\textwidth}
        \centering
        \includegraphics[width=\textwidth]{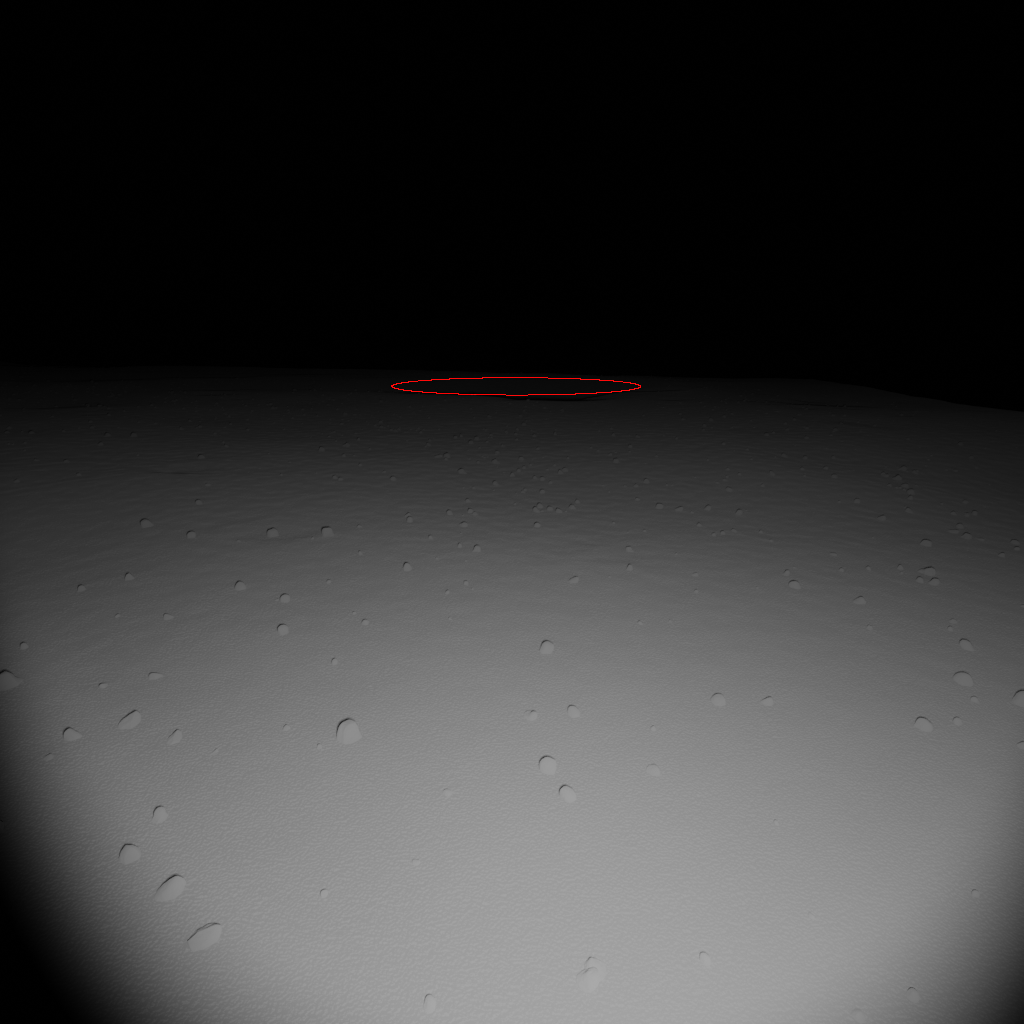}
        \caption{Ground Truth at \SI{17}{\meter}}
        \label{fig:qual_dets_c}
    \end{subfigure}
    \begin{subfigure}[t]{0.32\textwidth}
        \includegraphics[width=\textwidth]{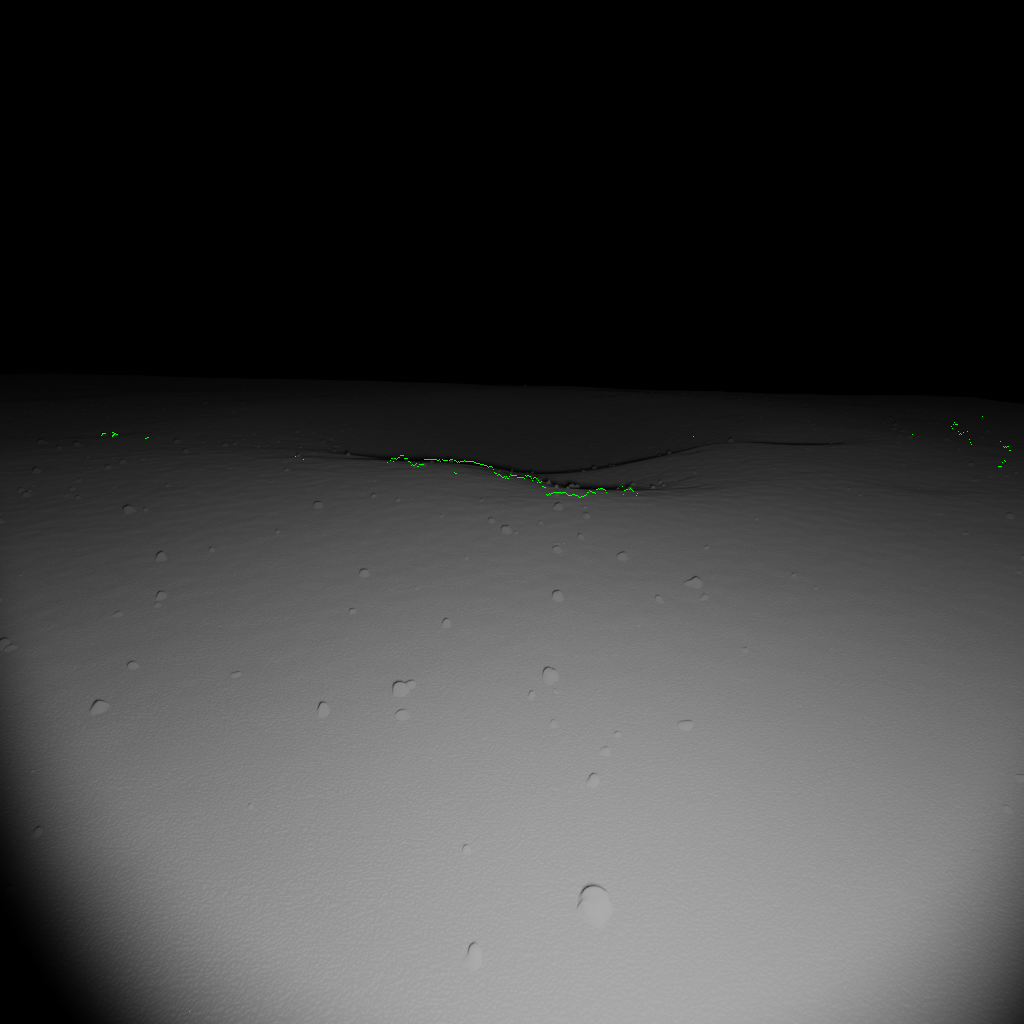}
        \caption{JPLV Disparity + Canny at \SI{7}{\meter}}
    \end{subfigure}
    \begin{subfigure}[t]{0.32\textwidth}
        \centering
        \includegraphics[width=\textwidth]{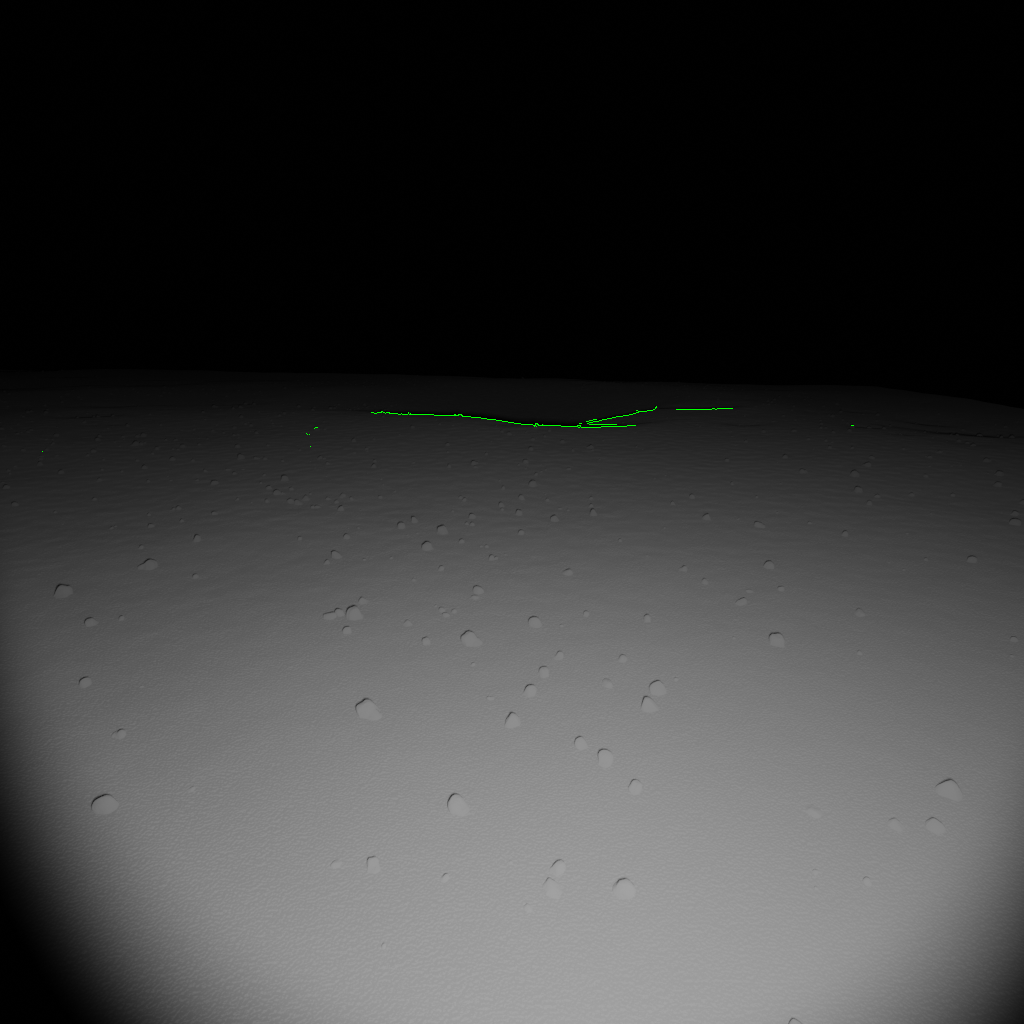}
        \caption{JPLV Disparity + Canny at \SI{12}{\meter}}
    \end{subfigure}
    \begin{subfigure}[t]{0.32\textwidth}
        \centering
        \includegraphics[width=\textwidth]{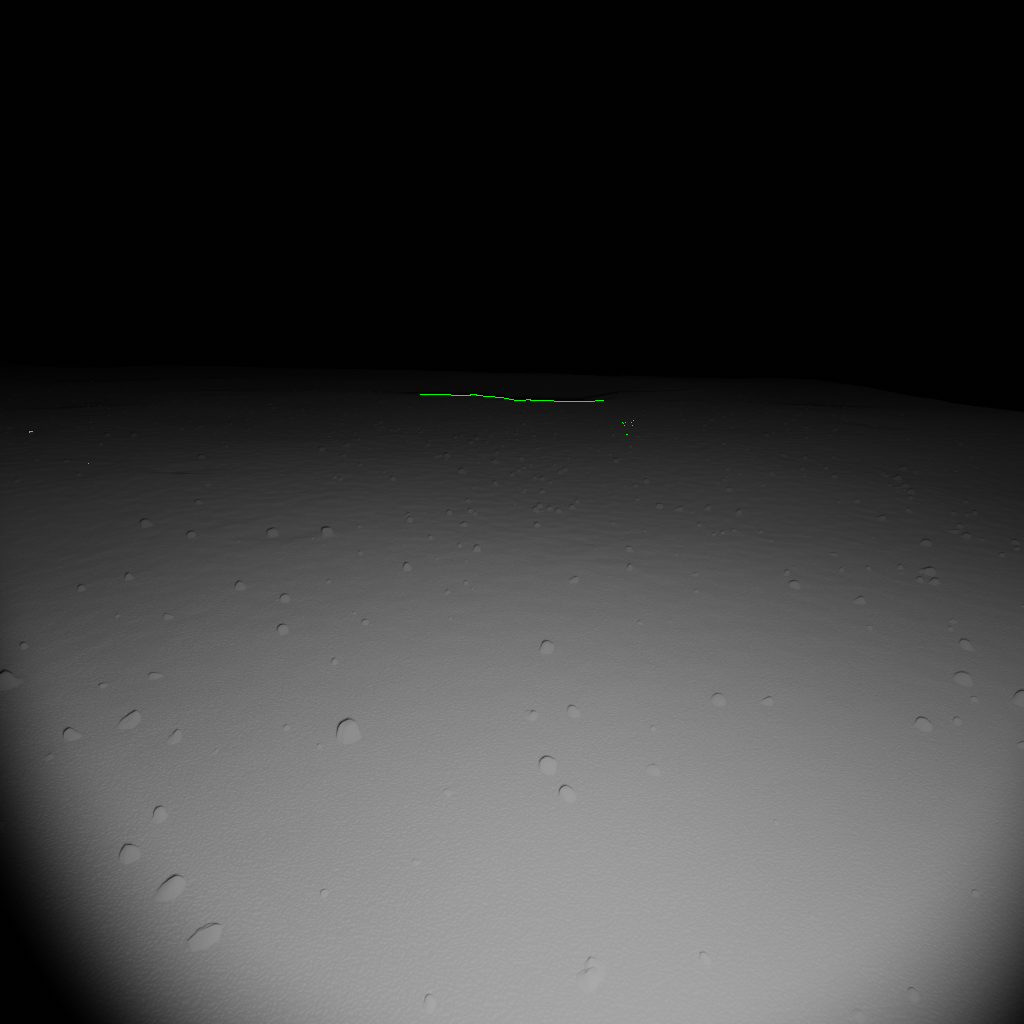}
        \caption{JPLV Disparity + Canny at \SI{17}{\meter}}
    \end{subfigure}
    \begin{subfigure}[t]{0.32\textwidth}
        \includegraphics[width=\textwidth]{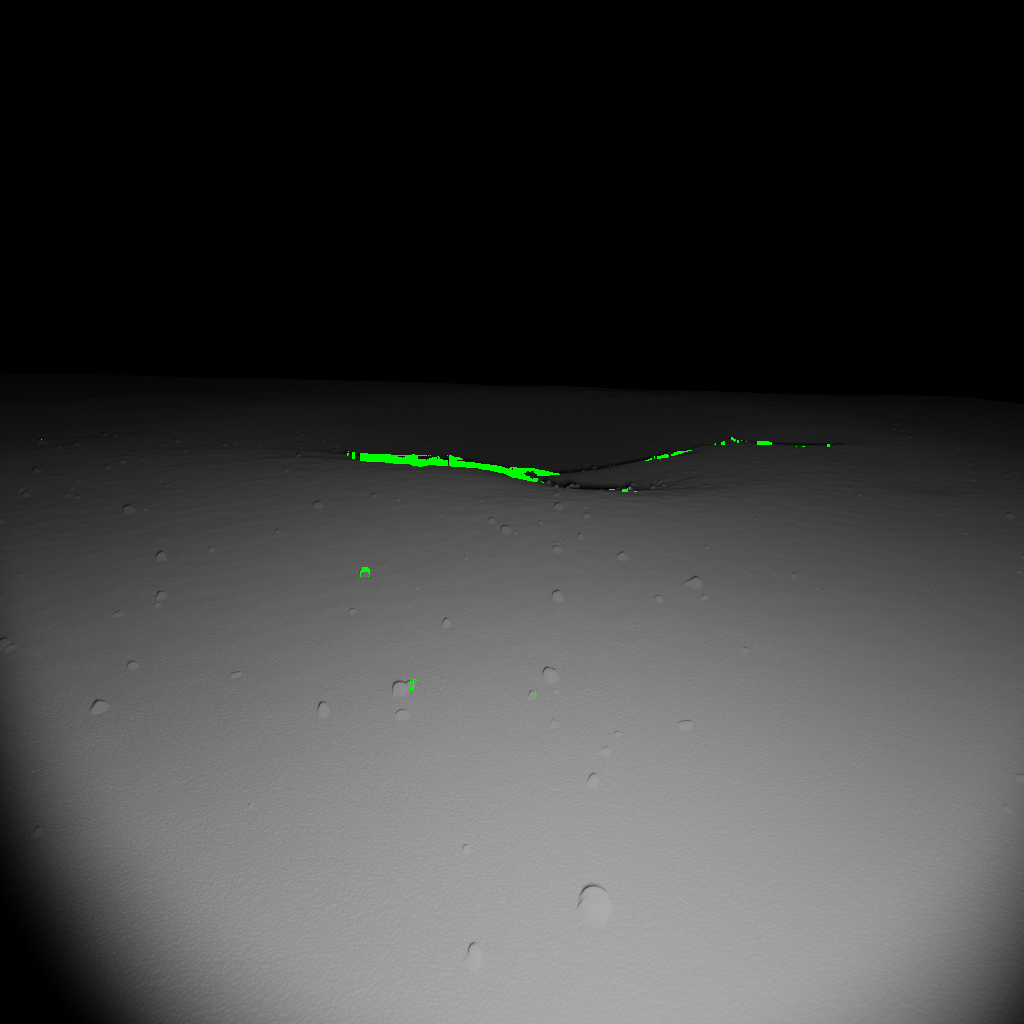}
        \caption{JPLV~\ac{hed} at \SI{7}{\meter}}
    \end{subfigure}
    \begin{subfigure}[t]{0.32\textwidth}
        \centering
        \includegraphics[width=\textwidth]{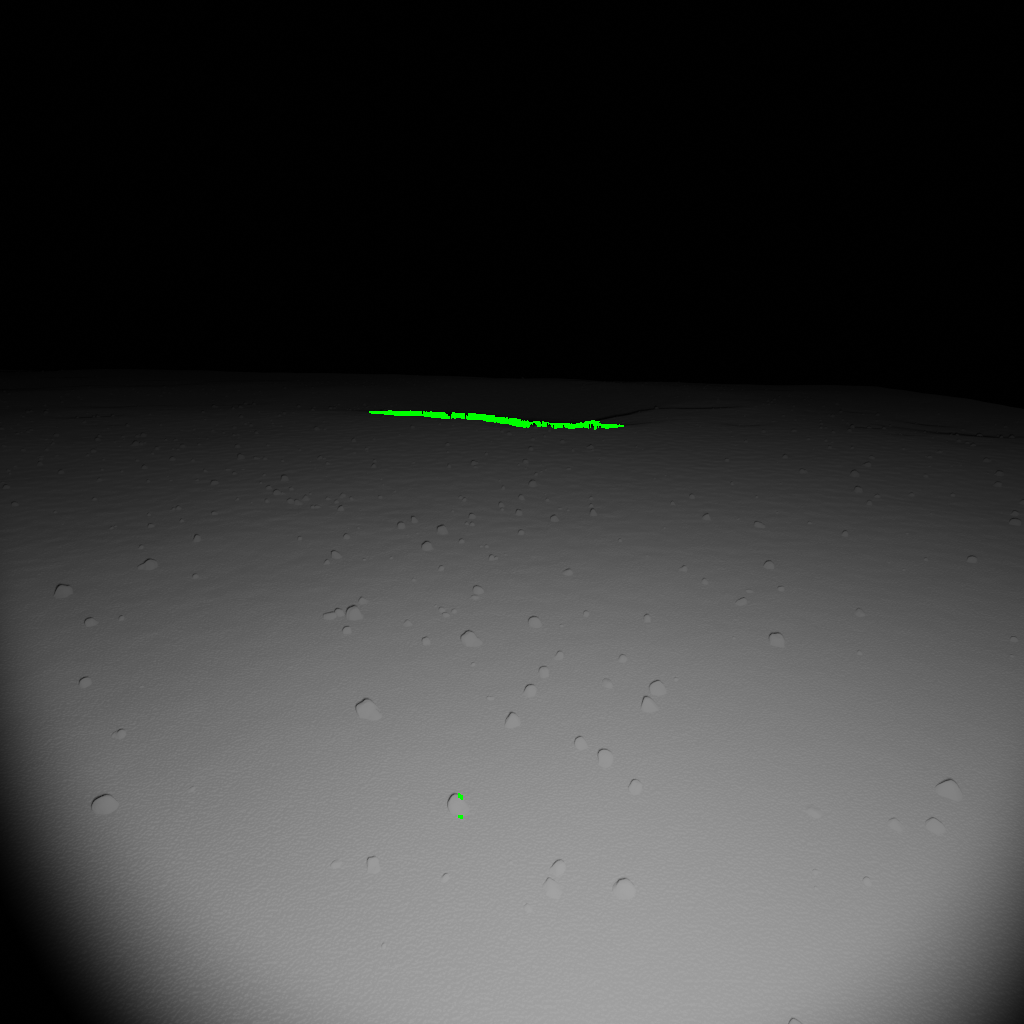}
        \caption{JPLV~\ac{hed} at \SI{12}{\meter}}
    \end{subfigure}
    \begin{subfigure}[t]{0.32\textwidth}
        \centering
        \includegraphics[width=\textwidth]{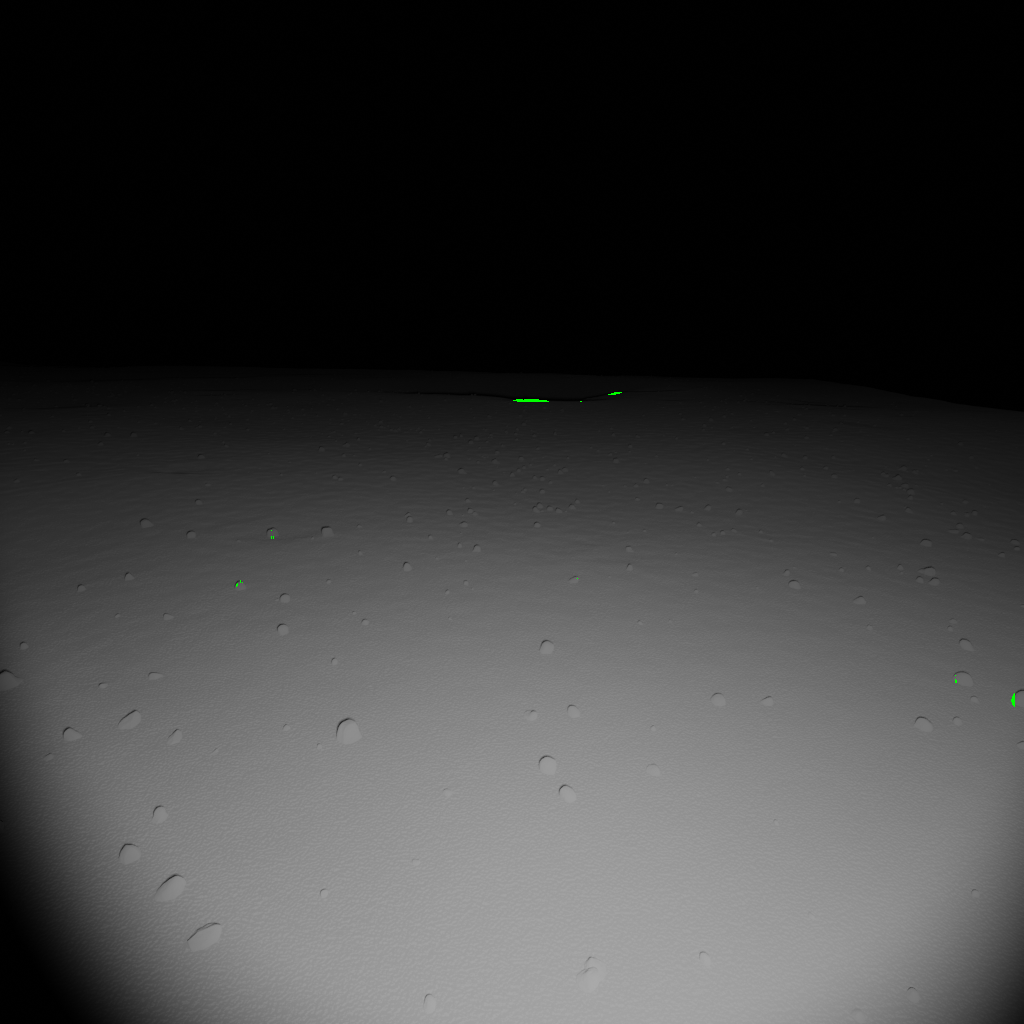}
        \caption{JPLV~\ac{hed} at \SI{17}{\meter}}
    \end{subfigure}
  \caption{\bf The efficacy of the JPLV~\ac{hed} approach over JPLV Disparity + Canny is demonstrated in simulations of crater rim detection overlay samples for crater 1.}
  \label{fig:qual_dets}
\end{figure*}

\begin{figure*}[t!]
    \centering
    \begin{subfigure}[t]{0.32\textwidth}
        \includegraphics[width=\textwidth]{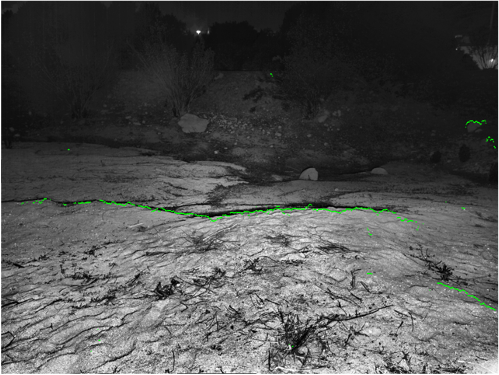}
        \caption{Negative obstacle \SI{5}{\meter} away}
    \end{subfigure}
    \begin{subfigure}[t]{0.32\textwidth}
        \centering
        \includegraphics[width=\textwidth]{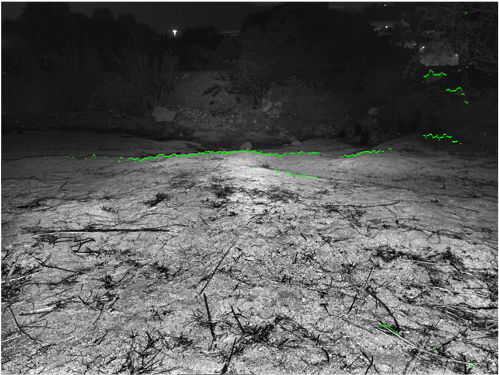}
        \caption{Negative obstacle \SI{10}{\meter} away}
    \end{subfigure}
    \begin{subfigure}[t]{0.32\textwidth}
        \centering
        \includegraphics[width=\textwidth]{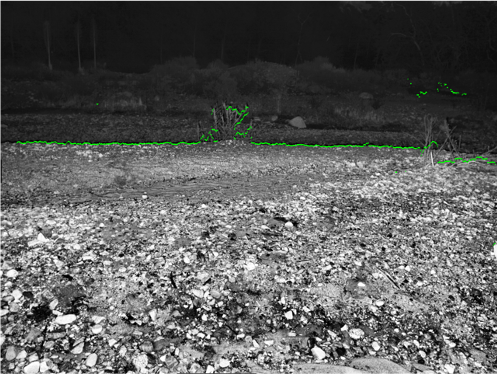}
        \caption{Negative obstacle \SI{10}{\meter} away}
    \end{subfigure}
    \begin{subfigure}[t]{0.32\textwidth}
        \includegraphics[width=\textwidth]{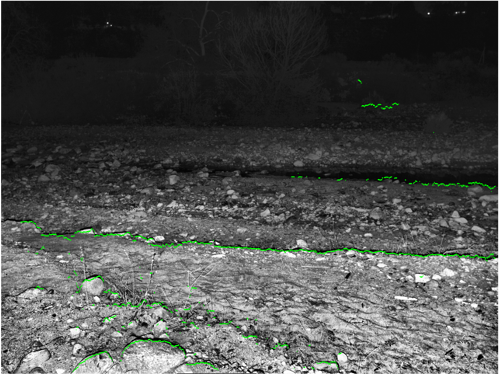}
        \caption{Negative obstacle \SI{5}{\meter} away}
    \end{subfigure}
    \begin{subfigure}[t]{0.32\textwidth}
        \centering
        \includegraphics[width=\textwidth]{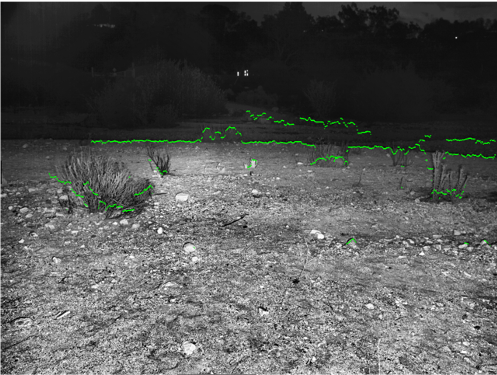}
        \caption{Negative obstacle \SI{15}{\meter} away}
    \end{subfigure}
    \begin{subfigure}[t]{0.32\textwidth}
        \centering
        \includegraphics[width=\textwidth]{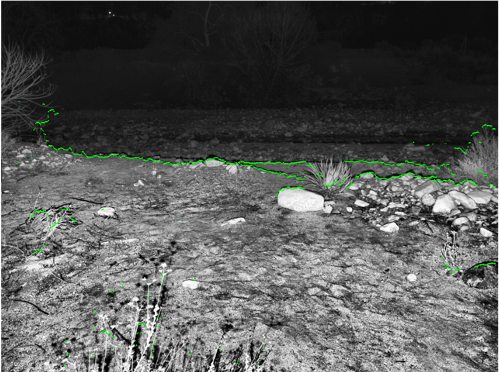}
        \caption{Negative obstacle \SI{10}{\meter} away}
    \end{subfigure}
  \caption{\bf Qualitative edge detections using JPLV disparity discontinuity detection and Canny hybrid on negative obstacles on a real dataset collected in a dry river bed at night. These results demonstrate the transferability of the crater detection algorithms from simulated data to a real environment.}
  \label{fig:arroyo_qual_dets}
\end{figure*}

\subsection{A. Metrics}

In order to evaluate the performance of surface crater detection, the dataset referenced in Section~\ref{sec:crater_dataset} was utilized.
Five different combinations of algorithms were evaluated.
These were disparity discontinuity detection within~\ac{sgbm} stereo, disparity discontinuity detection within JPLV stereo,~\ac{hed} using~\ac{sgbm} stereo,~\ac{hed} using JPLV stereo, and a hybrid JPLV disparity discontinuity detection and canny edge detection approach.
The hybrid discontinuity detection and canny approach was implemented so that Canny only ran on the portion of the image that was \SI{10}{\meter} away or further.
This was done since it was observed the discontinuity detection worked well in the near range but stereo began to degrade beyond \SI{10}{\meter}.

These algorithms were evaluated with two different metrics.
The first was an image based edge scoring method which captures an average Gaussian probability that a detected edge is on a ground truth crater edge. 
It utilizes a distance error computed in image space as represented in Equation \ref{eq:rover_q_score} where $\mathrm{Error}_{\mathrm{dist}_{px}}$ is the pixel error from ground truth to detection, $\mathrm{range}_{\mathrm{gt}}$ is the known ground truth range, $\mathrm{fl}$ is the focal length of camera, $\mathrm{ss}$ is the sensor size of the camera, and $\mathrm{Error}_{\mathrm{dist}}$ is the error in meters of the detection.
\begin{equation}
\mathrm{Error}_{\mathrm{dist}} = \mathrm{Error}_{\mathrm{dist}_{px}} * \frac{\mathrm{range}_{\mathrm{gt}}}{(\mathrm{fl} * \mathrm{ss})}~\label{eq:rover_q_score}
\end{equation}

The distance error was then passed into a Gaussian.
The Gaussian probabilities for all of the detected pixels were summed together and normalized by number of detected points to obtain a score.
This scoring method infused ground truth range values to remove the impact of stereo holes and stereo range uncertainty on the projection in order to better isolate the specific performance of the crater detection algorithms.
The sigma value for the Gaussian that was used in these experiments was $0.25\textrm{m}$.
This was chosen because the resolution of the~\ac{dem} utilized was \SI{0.25}{\meter}.
Therefore most detections should fall within this boundary if they are highly accurate.
The second metric used was "percent of front arc detected".
In this metric, there is a ground truth circle of the orbital crater.
Depending on the pose of the simulated cameras, the half arc of the ground truth circle that was nearest the simulated camera was projected into image space.
The crater detection was then matched to the half arc and the percentage of the half arc that was successfully identified was determined.
This metric removes the Gaussian component from the first metric; however, it does not capture false positives like the first metric.

\subsection{B. Detection Results on Simulated Data}

The results of running the different algorithms on the simulated dataset are observed in Figure \ref{fig:crater_det}.
There were several notable observations from the results.
The first was that the algorithms tended to perform the best around \SI{10}{\meter} and did not improve as craters came closer.
This was believed to be because as the camera gets closer to the crater, more of the crater becomes visible and the discontinuities become smaller.
However, as the crater becomes further than \SI{10}{\meter}, the stereo began to degrade.
Additionally, for the hybrid stereo and Canny technique, the Canny detection started detection at \SI{10}{\meter} and led to a significant jump in performance.

In terms of algorithm comparison, JPLV disparity discontinuity performed better than~\ac{sgbm} disparity discontinuity which is likely due JPLV having more holes than~\ac{sgbm}.
These holes at the boundary helped the disparity discontinuity detector find a better edge.
However, for~\ac{hed}, it performed well with either stereo technique, likely due to its representation of depth containing height values.
\ac{hed} was used with its out of the box weights from its authors.
It likely could be improved with finetuning on a Lunar dataset. 

In addition to quantitative results, samples of crater rim detection overlays are in Figure \ref{fig:qual_dets}.
These results were on crater 1 which is a nearly \SI{10}{\meter} in diameter crater.
Both methods were able to detect the craters well, but JPLV~\ac{hed} did have more falloff at $\SI{17}{\meter}$ than the Canny detector.
However, the Canny edge detector was optimized for this environment where as~\ac{hed} was a generalized detector.
Overall the generalization of~\ac{hed} was extremely promising as a crater rim detection approach.

\subsection{C. Detection Results on Real Data}

As described previously, data was collected from a location with negative obstacles at night. 
This dataset was used to validate the performance of stereo and crater detection algorithms. 
Figure \ref{fig:arroyo_stereo} presents a sample of \SI{5}{\meter} and \SI{10}{\meter} negative obstacles and the corresponding stereo results from JPLV. From this figure is was observed that stereo is dense up unto the leading edge of the negative obstacle.
Additionally, at \SI{5}{\meter}, the far edge of the negative obstacle was captured in the disparity values. At \SI{10}{\meter}, the far edge, did contain some disparity values but it was sparse. 
While not fully representative of the Lunar surface, this demonstrated that current stereo techniques do have the capability to work in low light conditions at the ranges necessary.
The data was also used to evaluate the edge detection techniques. 
The JPLV disparity discontinuity and Canny edge detection hybrid was found to be the best on simulation data and therefore it was used on the real data. 
Figure \ref{fig:arroyo_qual_dets} demonstrates sample detections at different ranges.
These detection results did contain false positives on some of the vegetation as the false positive rejection was not run. 
Vegetation is not present on the moon, however, objects such as rocks could present similar issues.
Overall, the negative obstacle edge detection qualitatively performs well.

\begin{figure*}
    \centering
    \begin{subfigure}[t]{0.23\textwidth}
        \includegraphics[width=\textwidth]{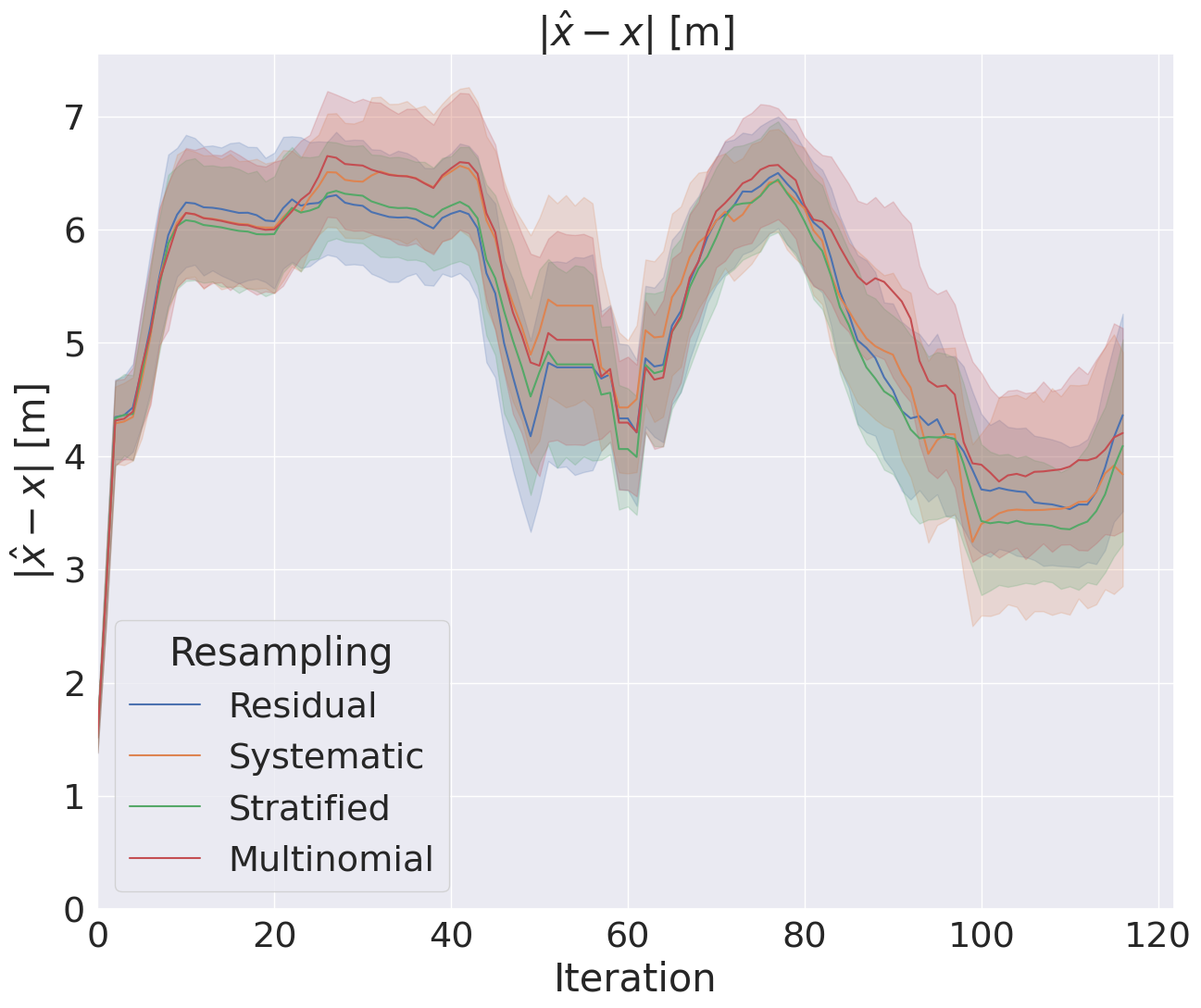}
        \caption{Ground truth error for traj. 1.}
        \label{fig:sim_2_traj_1_gt_error}
    \end{subfigure}
    \begin{subfigure}[t]{0.23\textwidth}
        \includegraphics[width=\textwidth]{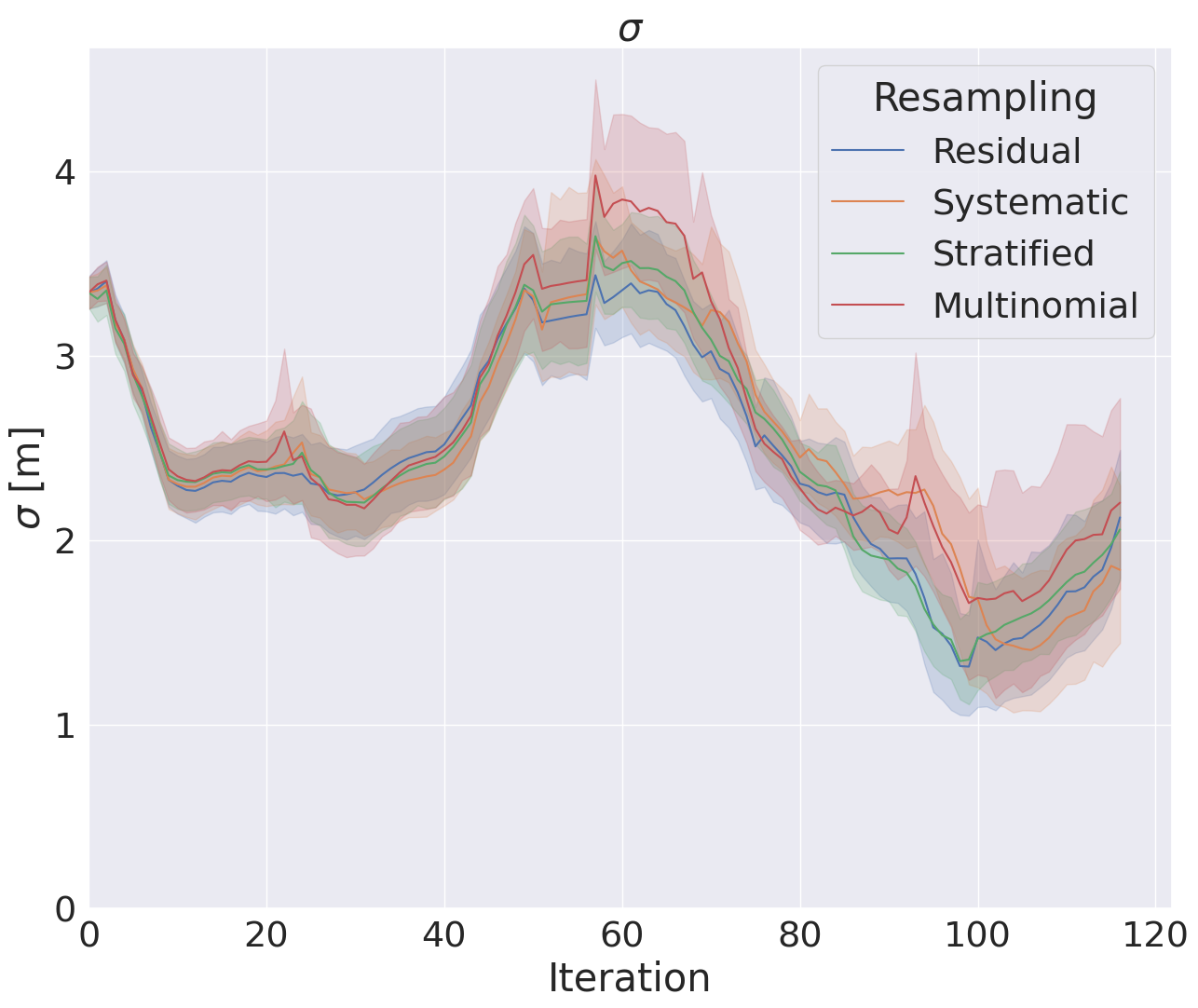}
        \caption{Filter covariance for traj. 1.}
        \label{fig:sim_2_traj_1_cov}
    \end{subfigure}
    \begin{subfigure}[t]{0.23\textwidth}
        \includegraphics[width=\textwidth]{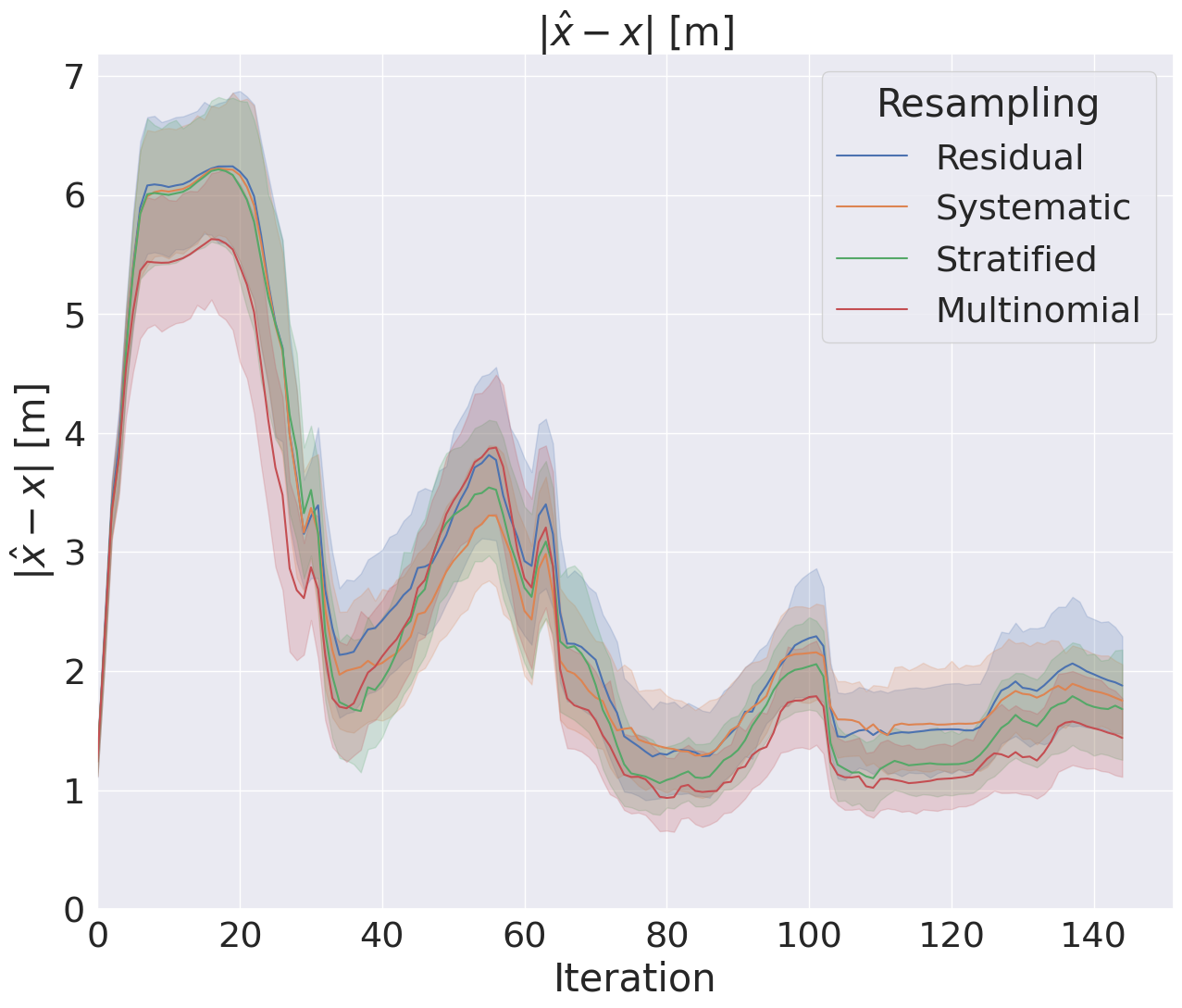}
        \caption{Ground truth error for traj. 2.}
        \label{fig:sim_2_traj_diag_gt_error}
    \end{subfigure}
    \begin{subfigure}[t]{0.23\textwidth}
        \includegraphics[width=\textwidth]{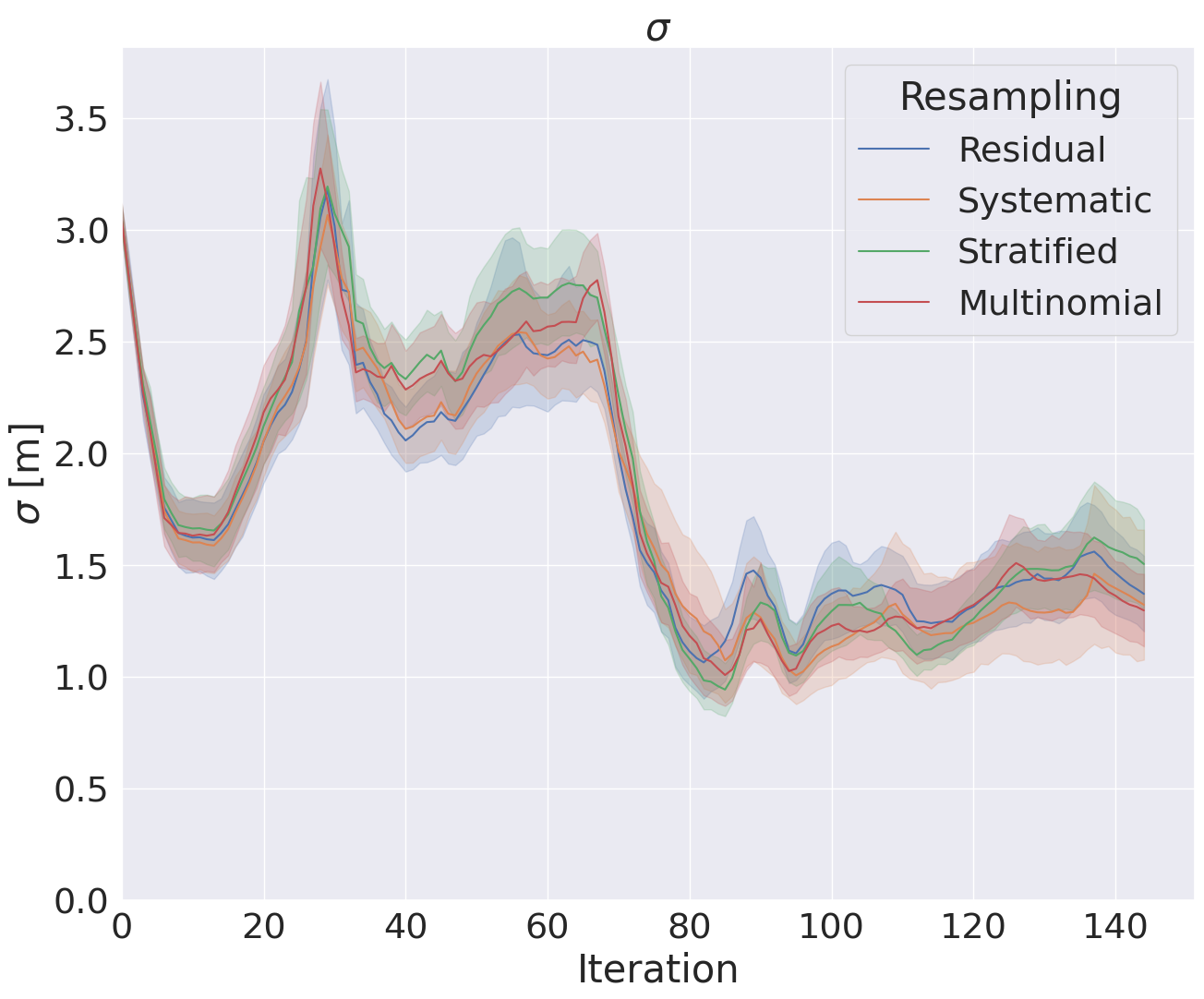}
        \caption{Filter covariance for traj. 2.}
        \label{fig:sim_2_traj_diag_cov}
    \end{subfigure}
  \caption{\bf A comparison of the four proposed resampling schemes demonstrated that systematic resampling empirically outperforms the other scheme in terms of relatively lower ground truth error and reduced uncertainty in the filter.}
  \label{fig:sim_2_resampling}
\end{figure*}

\begin{figure}[t!]
    \centering
    \hfill
    \begin{subfigure}[t]{0.23\textwidth}
        \centering
        \includegraphics[width=\textwidth]{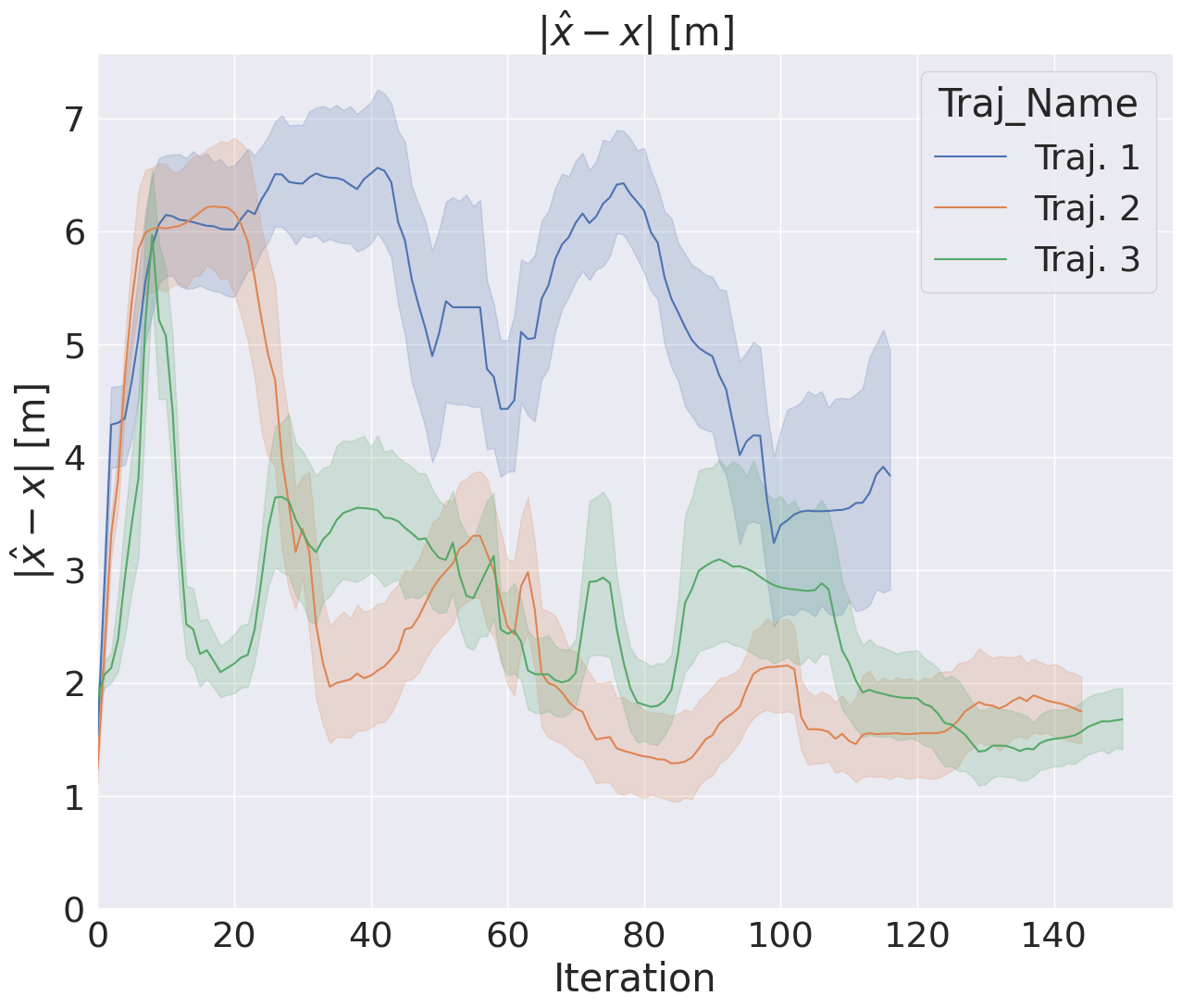}
        \caption{Ground truth error.}
        \label{fig:mc_traj_gt_err}
    \end{subfigure}
    \hfill
    \begin{subfigure}[t]{0.23\textwidth}
        \centering
        \includegraphics[width=\textwidth]{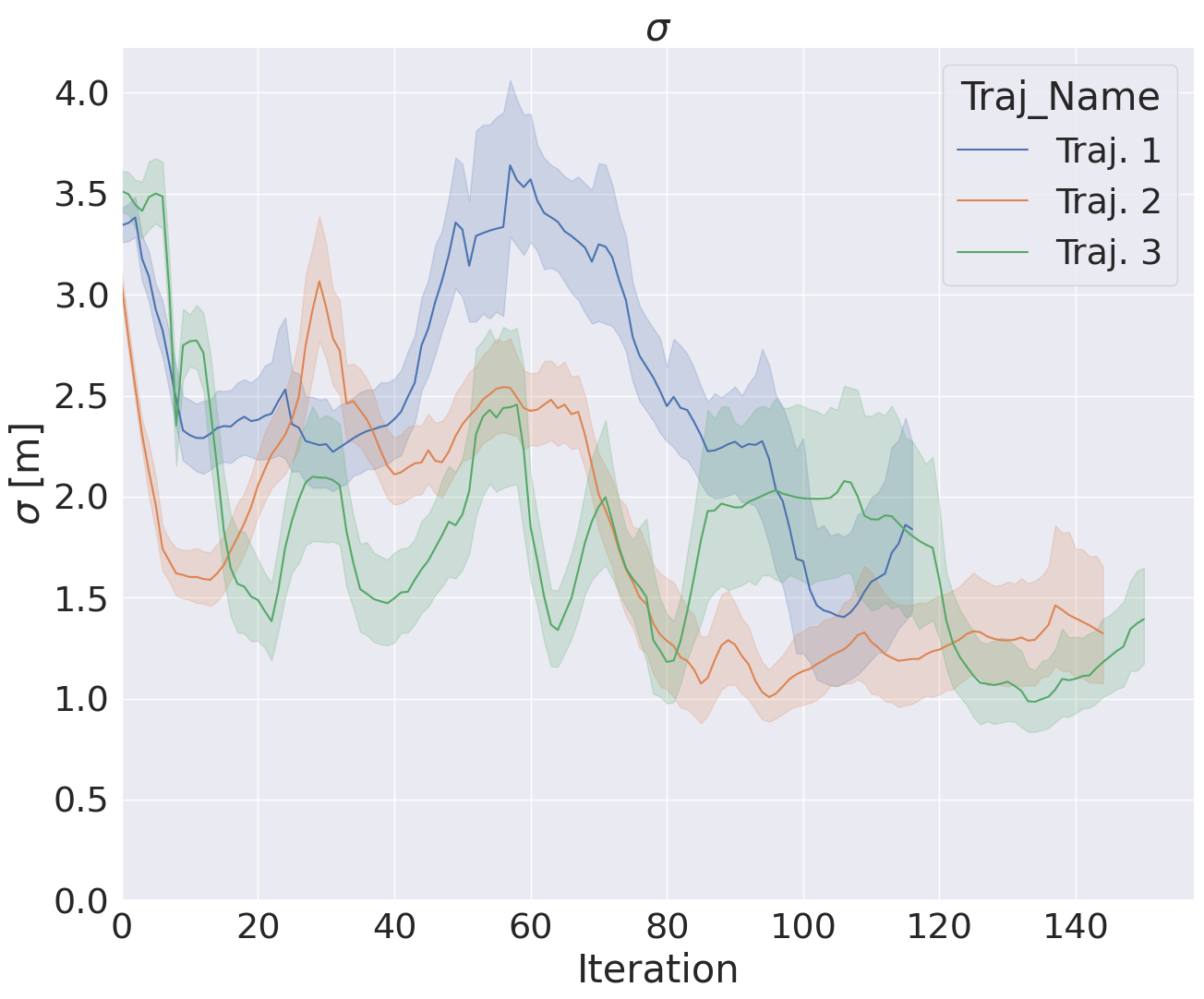}
        \caption{Filter covariance.}
        \vspace{-10pt}
        \label{fig:mc_traj_cov}
    \end{subfigure}
    \hfill
  \caption{\bf Monte Carlo simulations for trajectories 1--3 demonstrated the efficacy of the Q-Score based particle filtering approach at accomplishing global rover localization.}
\label{fig:mc_sim_traj}
\end{figure}

\begin{figure}
    \centering
    \hfill
    \begin{subfigure}[t]{0.23\textwidth}
        \centering
        \includegraphics[width=\textwidth]{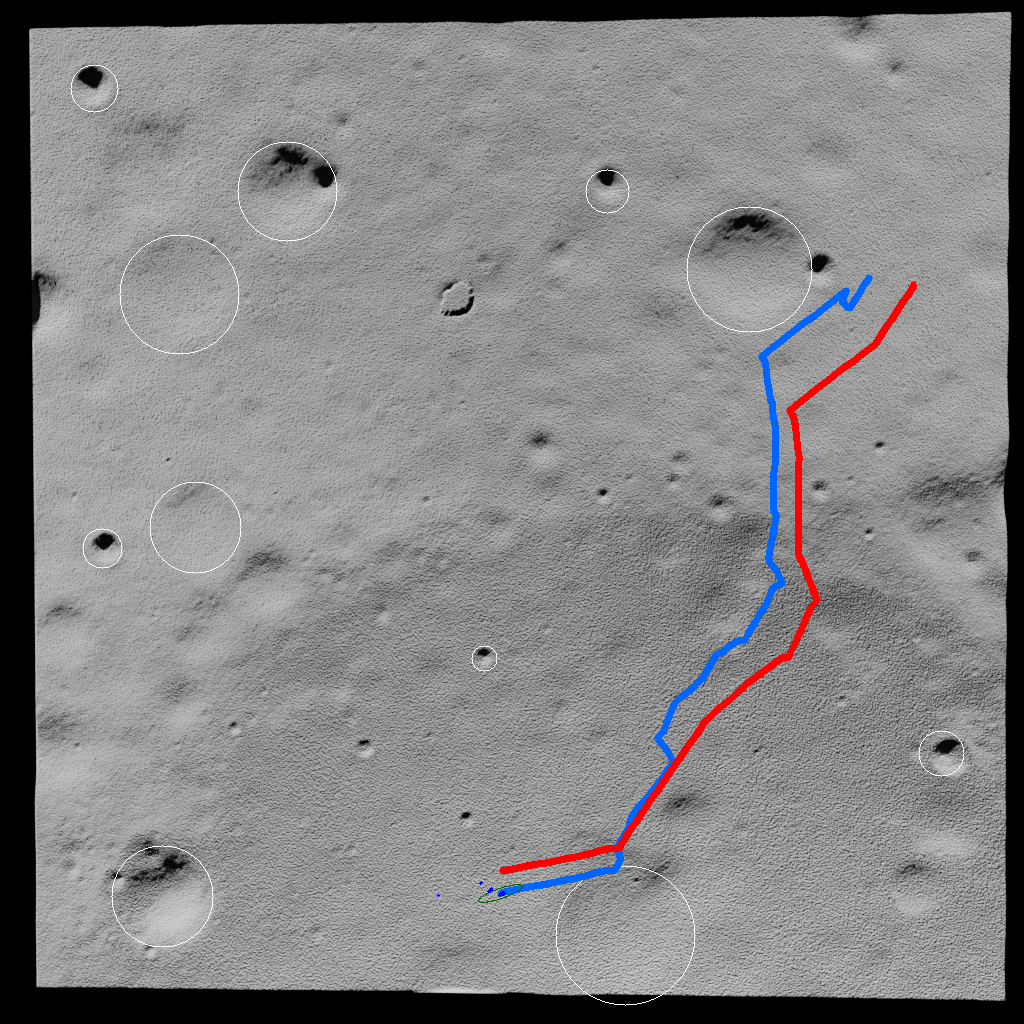}
        \caption{Traj. 1 traverse -- case A.}
        \label{fig:mc_traj_1_case_A}
    \end{subfigure}
    \hfill
    \begin{subfigure}[t]{0.23\textwidth}
        \centering
        \includegraphics[width=\textwidth]{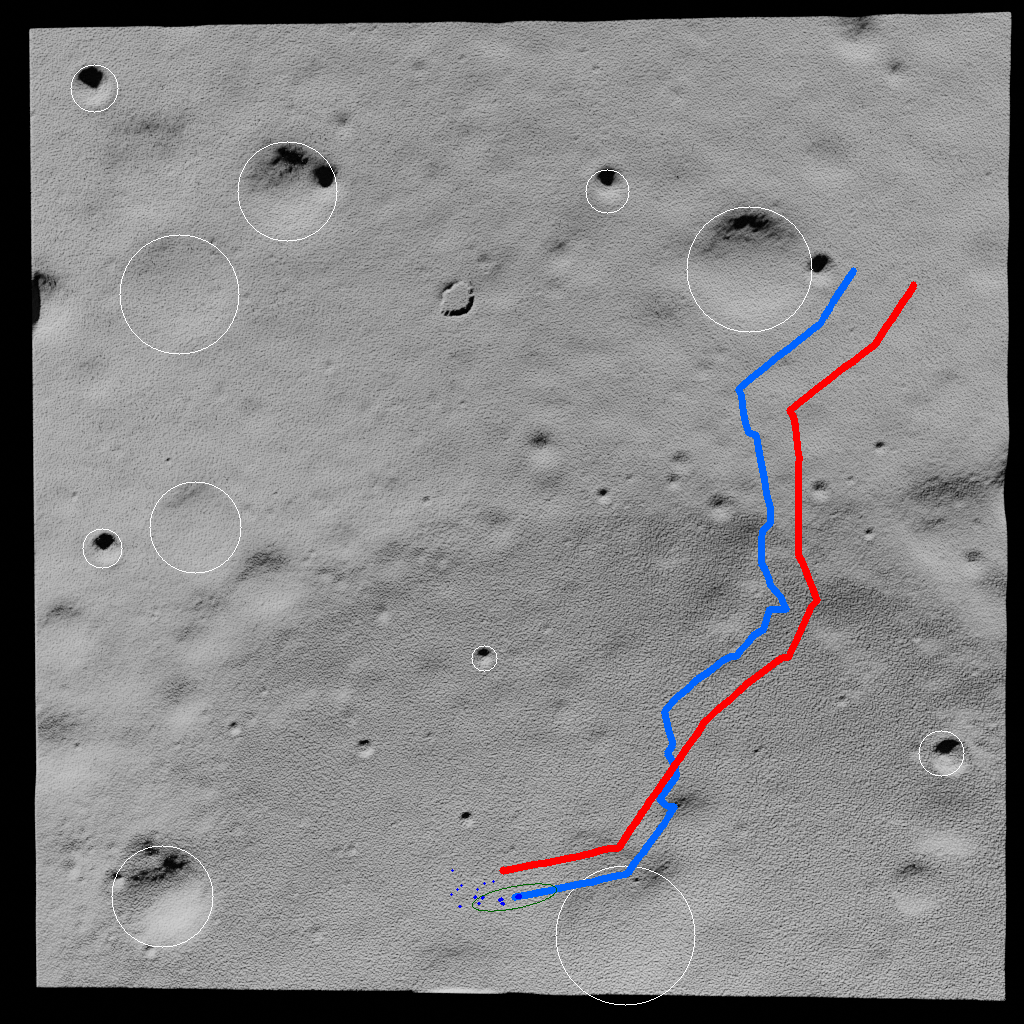}
        \caption{Traj. 1 traverse -- case B.}
        \label{fig:mc_traj_1_case_B}
    \end{subfigure}
    \hfill
  \caption{\bf Two Monte Carlo trials for trajectory 1 are illustrated with the ground truth in red and the weighted average belief $\mu^t$ in blue. The comparatively better performance of the filter in case A {\em (left)} was due to false positive crater rim measurements in case B {\em (right)} that led to worse localization.}
\label{fig:mc_sim_traj_1_cases}
\vspace{-10pt}
\end{figure}

\section{Localization Performance}
In this section, we provide Monte Carlo results on the performance of the proposed ShadowNav filtering algorithm.
For each simulation, we analyzed the performance of the ShadowNav filter on the basis of the following metrics:

{\em Ground truth error:} We computed the weighted average mean $\mu^t = \sum_{i=1}^{N_s} w_i^t b_i^t$ at time $t$ for the filter using the particle weights and beliefs and compute the $\ell_2$-distance to the ground truth ${\textrm{gt}}^t$, i.e., $\|\mu^t - {\textrm{gt}}^t\|_{2}$.

{\em Particle filter uncertainty:} To capture the uncertainty associated with the current belief, we additionally computed the weighted covariance matrix $\Sigma^t = \sum_{i=1}^{N_s}\tilde{w}_i^t (b_i^t - \mu^t)(b_i^t-\mu^t)$, where $\tilde{w}_i^t$ are the normalized weights detailed in Alg.~\ref{alg:systematic_resampling}.
The metric we report at each time step was the square root of the largest eigenvalue $\sqrt{\lambda_\textrm{max} (\Sigma^t)}$, which corresponded to the worst case variance of the estimation error~\cite{JoshiBoyd2009,CarloneKaraman2019}.

{\em Mahalanobis distance:} The final metric we computed was the Mahalanobis distance, which measures the distance between and the particle filter distribution and ground truth position.
We approximately computed this by fitting a Gaussian distribution ${\mathcal{N}}(\mu^t, \Sigma^t)$ to the particle filter distribution, for which the Mahalnobis distance is simply a weighted $\ell_2$-norm $\sqrt{(\mu^t - {\textrm{gt}}^t)^T(\Sigma^t)^{-1}(\mu^t - {\textrm{gt}}^t)}$.

\subsection{A. Resampling Scheme Comparison}
In this section, we compared the baseline systematic resampling approach detailed in Alg.~\ref{alg:systematic_resampling} against three other resampling methods utilized: multinomial, residual, and stratified (we refer the reader to~\cite{ArulampalamMaskellEtAl2002,LiBolicEtAl2015,KuptameteeAunsri2022} for a thorough review of these approaches.)
Figure~\ref{fig:sim_2_resampling} presents the ground truth error and filter uncertainty for the four different resampling approaches.
We saw that, for the two trajectories compared in Figure~\ref{fig:sim_2_resampling}, systematic resampling led to comparable ground truth error as the other resampling approaches, but that systematic resampling outperformed the other approaches in terms of the overall uncertainty of the filter.
Indeed, we note that multinomial resampling, the most commonly employed resampling technique, fared quite poorly in terms of the variance of the filter uncertainty (Fig.~\ref{fig:sim_2_traj_1_cov} and~\ref{fig:sim_2_traj_diag_cov}).


\subsection{B. Baseline Performance Evaluation}
Finally, we evaluated the performance of the proposed ShadowNav particle filter approach on three test trajectories.
Our analysis consisted of Monte Carlos simulations with 25 seeds and utilizing 2\% odometry noise and initial belief distribution with $\sigma_0=$\SI{3}{\meter}.
Each simulation was run with $N_s=100$ particles and systematic resampling as the resampling scheme with $N_{\textrm{eff,thresh}}=50$ as the resampling threshold.

Figure~\ref{fig:mc_sim_traj} shows Monte Carlo simulation results for the three test trajectories.
We saw that the initial uncertainty in the filter began at approximately \SI{3}{\meter} as expected by sampling from a distribution with $\sigma_0=$\SI{3}{\meter}.
Thereafter, the filter was able to improve the rover position estimate, which led to an absolute error reduction of \SI{4}{\meter}.
Further, we see in Table~\ref{table:mc_final_values} that the metrics computed at the final time step indicate convergence of the filter, with an average final error of $\leq$\SI{4}{\meter} and an absolute error reduction of \SI{4}{\meter}.

\begin{figure*}
    \centering
    \hfill
    \begin{subfigure}[t]{0.3\textwidth}
        \centering
        \includegraphics[width=\textwidth]{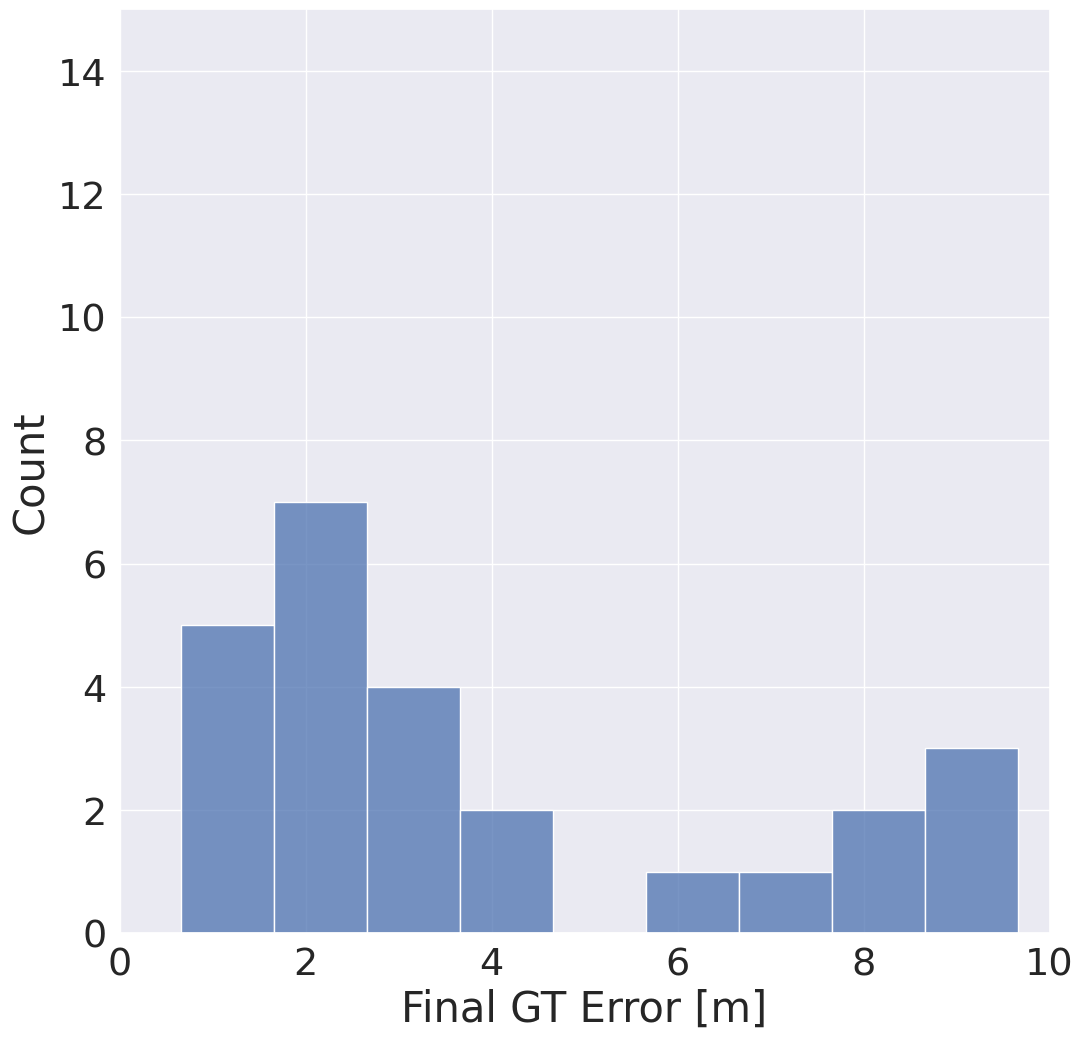}
        \caption{Trajectory 1}
        \label{fig:mc_traj_1_gt_error_hist}
    \end{subfigure}
    \hfill
    \begin{subfigure}[t]{0.3\textwidth}
        \centering
        \includegraphics[width=\textwidth]{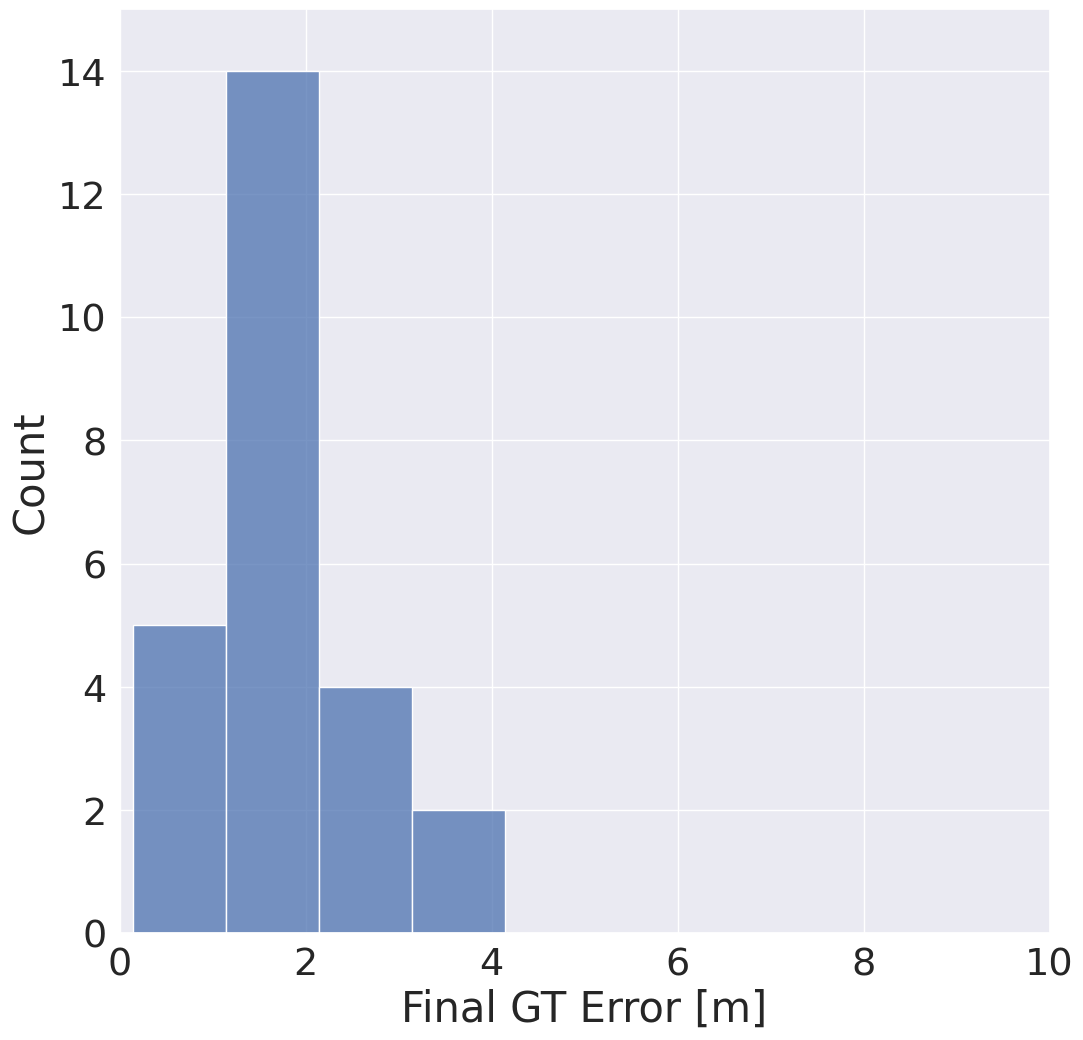}
        \caption{Trajectory 2}
        \label{fig:mc_traj_2_gt_error_hist}
    \end{subfigure}
    \hfill
    \begin{subfigure}[t]{0.3\textwidth}
        \centering
        \includegraphics[width=\textwidth]{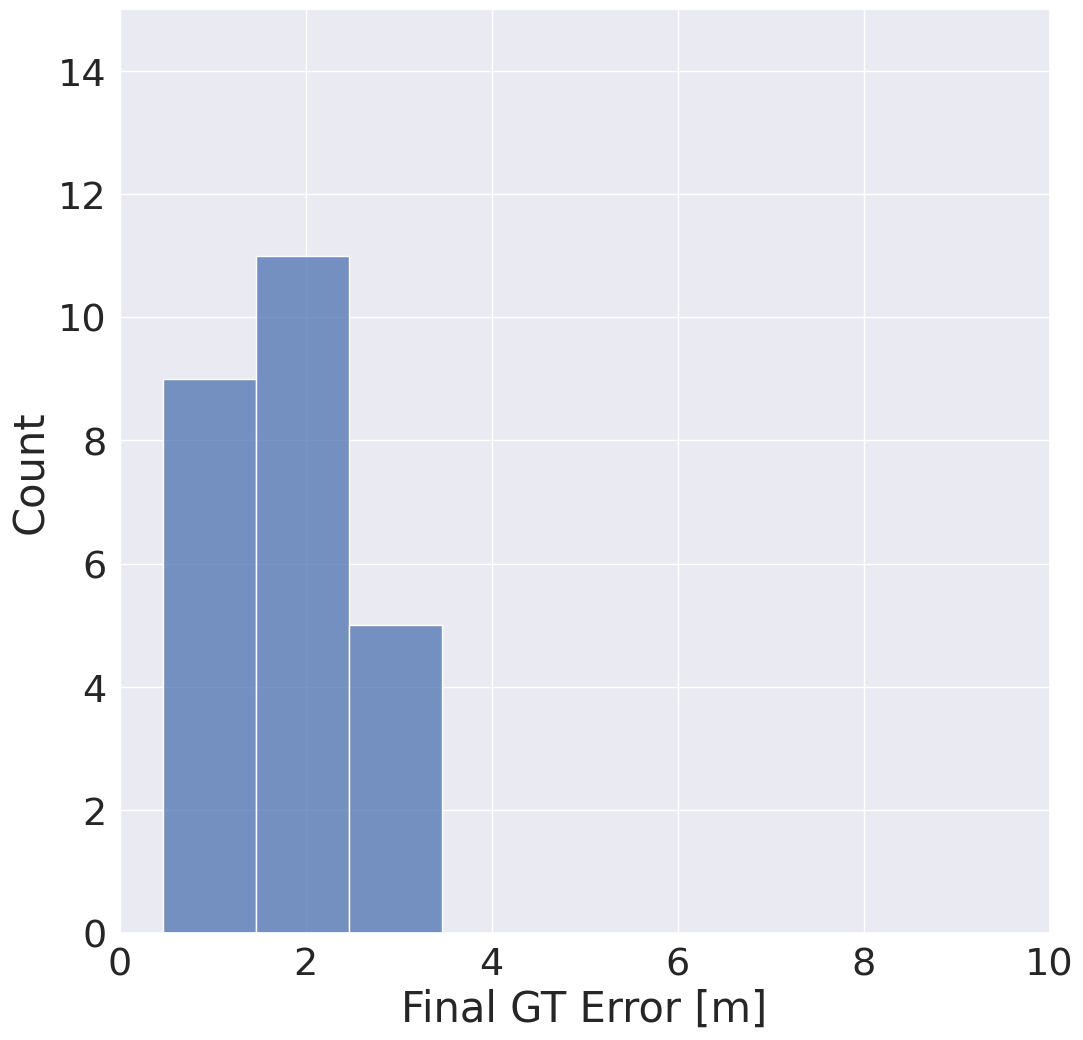}
        \caption{Trajectory 3}
        \label{fig:mc_traj_3_gt_error_hist}
    \end{subfigure}
    \hfill
  \caption{\bf The final ground truth error distribution for 25 Monte Carlo simulations showed filter convergence to $\leq4\textrm{m}$ error in all cases for trajectories 2 and 3 and for the majority of cases for trajectory 1.}
  \label{fig:mc_gt_error_hist}
\end{figure*}

\begin{table}
\renewcommand{\arraystretch}{1.3}
\caption{\bf The metrics computed at the end of a long-range lunar traverse indicate convergence of the particular filter on trajectories 2 and 3, but spurious measurements from unlabeled crater lead to relatively poor performance on trajectory 1.}
\label{table:mc_final_values}
\centering
\begin{tabular}{|l|c|c|c|}
\hline
& \bfseries Error & \bfseries Uncertainty & \bfseries Mahalanobis Dist. \\
\hline\hline
Traj. 1 & $3.84 \pm 2.78$ & $1.84 \pm 1.12$ & $8.74 \pm 10.03$ \\
Traj. 2 & $1.75 \pm 0.78$ & $1.32 \pm 0.76$ & $2.75 \pm 1.88$\\
Traj. 3 & $1.68 \pm 0.7$ & $1.39 \pm 0.61$ & $2.92 \pm 1.91$\\
\hline
\end{tabular}
\end{table}

As seen in Figure~\ref{fig:mc_gt_error_hist}, while the filter performed well on trajectories 2 and 3, the filter was less performant for the trajectory 1 test case.
Figure~\ref{fig:mc_sim_traj_1_cases} illustrates the performance of the filter on trajectory 1 for two different random seeds as the rover starts from the northern edge of the orbital map and moves southward.
During the middle portion of this traverse, the craters were out-of-sight for the rover and, as we see in Figure~\ref{fig:mc_sim_traj}, false positive observations led to increases in the error and uncertainty in the filter.
As the crater in the southern portion of the orbital map became observable for the rover, we saw that the estimate quickly improved in case A (Fig.~\ref{fig:mc_traj_1_case_A}), but continues to have a residual error in case B (Fig.~\ref{fig:mc_traj_1_case_B}).
This poor convergence behavior was also explained by false positive observations, wherein the filter had difficulty reconciling the front edge of the rim with the back edge, an issue that requires further investigation.

\subsection{C. Debugging}

\begin{figure}
    \centering
    \hfill
    \begin{subfigure}[t]{0.23\textwidth}
        \centering
        \includegraphics[width=\textwidth]{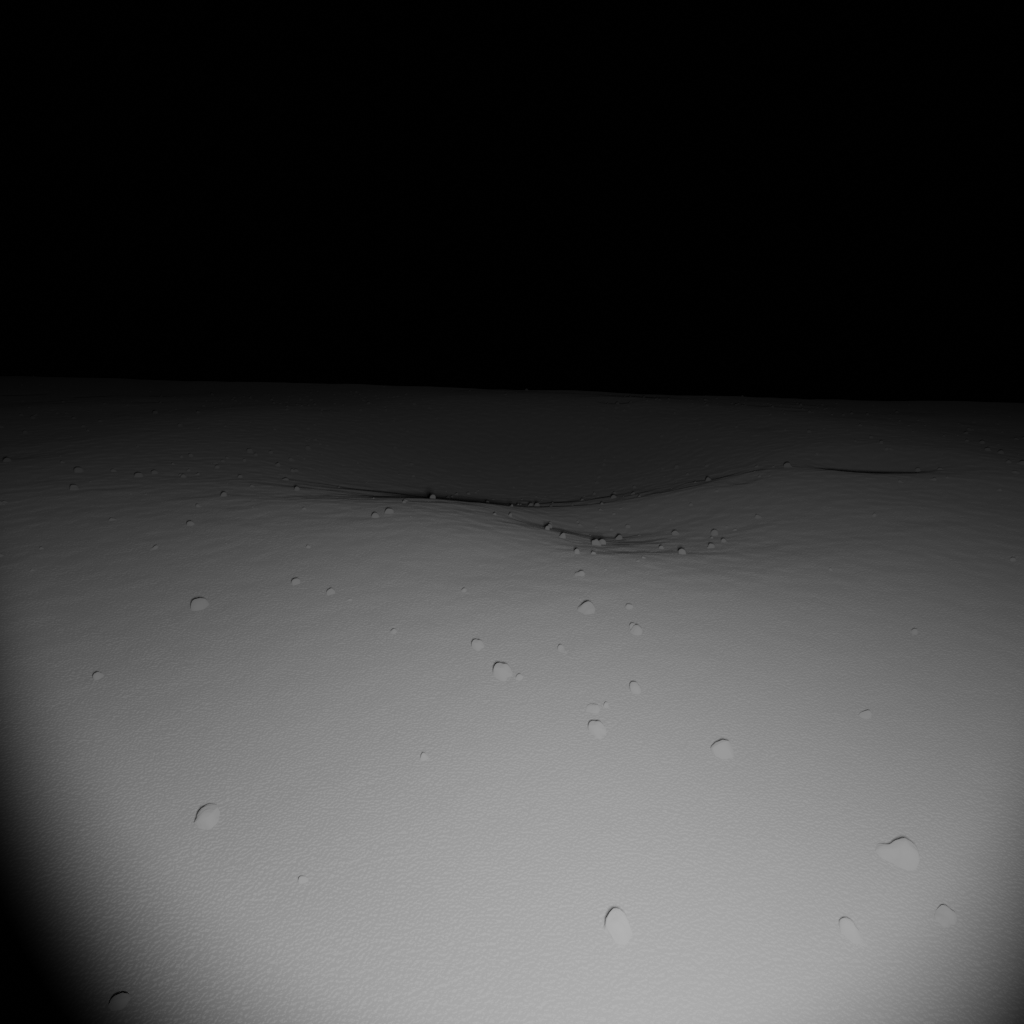}
        \vspace{-10pt}
        \caption{Simulated image from Blender.}
        \label{fig:crat1a}
    \end{subfigure}
    \hfill
    \begin{subfigure}[t]{0.23\textwidth}
        \centering
        \includegraphics[width=\textwidth]{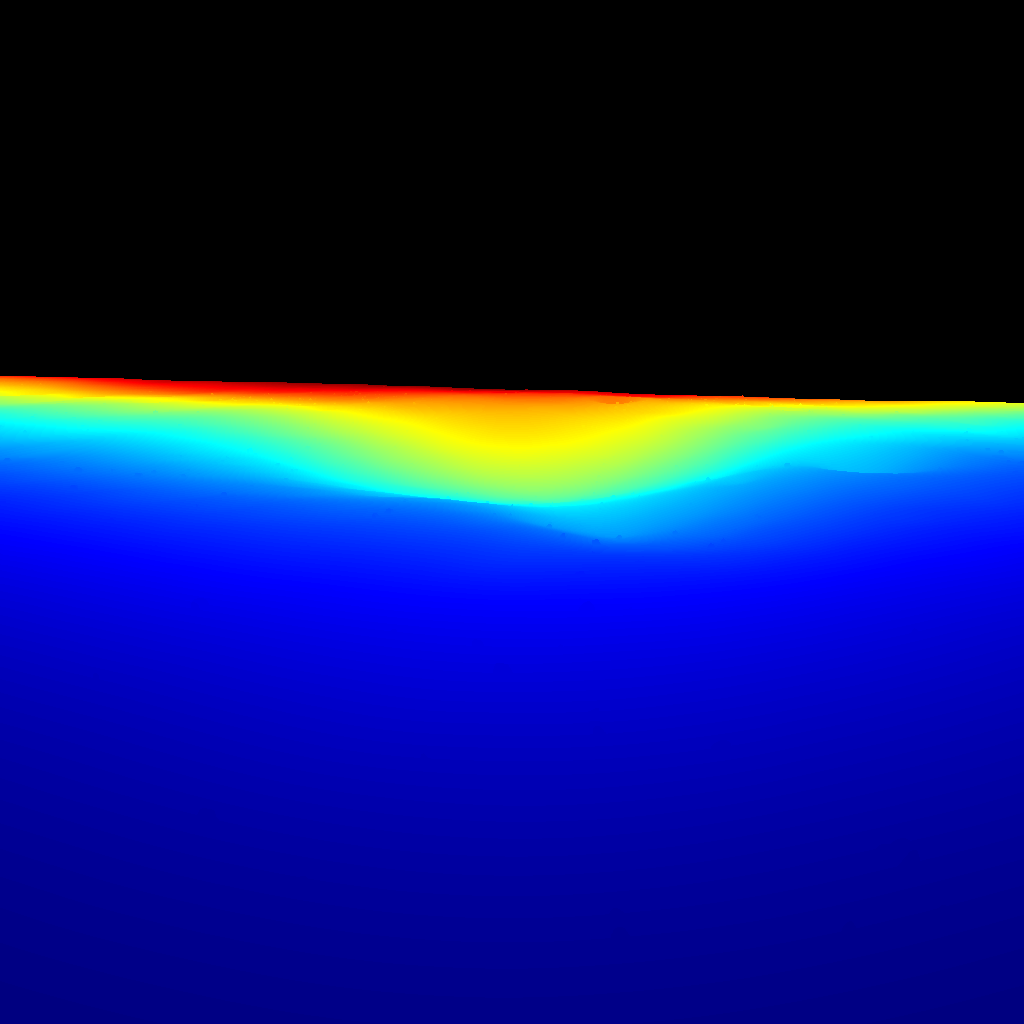}
        \caption{Perfect depth: blue is close, red is far}
        \vspace{-10pt}
        \label{fig:perfect-depth}
    \end{subfigure}
    \hfill
  \caption{\bf Crater 1 viewed from \SI{5}{\meter} away from front rim.}
\label{fig:crater1}
 \vspace{-10pt}
\end{figure}

When testing the particle filter, we found it helpful to generate ``perfect'' datasets where ground truth depth was generated directly from the simulator as shown in Figure~\ref{fig:perfect-depth}) and crater edges were plotted into the rover frame using their exact known world coordinates (see Fig.~\ref{fig:qual_dets_a}-\ref{fig:qual_dets_c}).
This approach uncovered bugs with our perception and projection pipeline as well as the particle filter pipeline and it is highly recommended to build such a dataset for all similar work.

\vspace{-5pt}
\section{Conclusions}
In this work we present a system to perform autonomous absolute localization on a Lunar rover while it is in darkness.
This system entails using a stereo camera and illuminator.
We enhanced a Blender based simulation with a custom Lunar texture and an implementation of the Hapke model to model surface reflectance as accurately as possible.
We further demonstrate both geometric and learning based techniques for detecting the leading edge of a crater with ability to detect some craters out to \SI{20}{\meter} range.
We propose a method of matching the detected leading crater rims with known craters within an orbital map and using these matches to score observations with our Q-Score.
Finally we demonstrate absolute localization within our simulation environment with less than \SI{4}{\meter} error, and an absolute error reduction of \SI{4}{\meter} upon detecting craters.
These results show promise for further investigation in the future on more simulation environments as well as on to be collected real analogue datasets.

\subsection{D. Future Work}
In the future, we seek to perform several updates and additional evaluations.
The primary focus is to experimentally collect a nighttime dataset using representative hardware in an analogue Lunar environment with negative obstacles to evaluate the system.
Additional evaluation is planned to evaluate the performance of the proposed approach along longer trajectories, on more varied Lunar type locales, and for different rover specific parameters such as camera height off of the ground.
Finally, we plan to validate our proposed approach on a flight-like embedded computer (\eg{}, a Snapdragon) to demonstrate that it is computationally feasible for use onboard a Lunar rover.
\vspace{-10pt}


\acknowledgments
The research was carried out at the Jet Propulsion Laboratory, California Institute of Technology, under a contract with the National Aeronautics and Space Administration (80NM0018D0004).
The authors would like to thank Yang Cheng, Olivier Lamarre, and Scott Tepsuporn for their discussions during the development of this work.


\bibliographystyle{IEEEtran} 
\bibliography{main}

\thebiography
\begin{biographywithpic}
{Abhishek Cauligi}{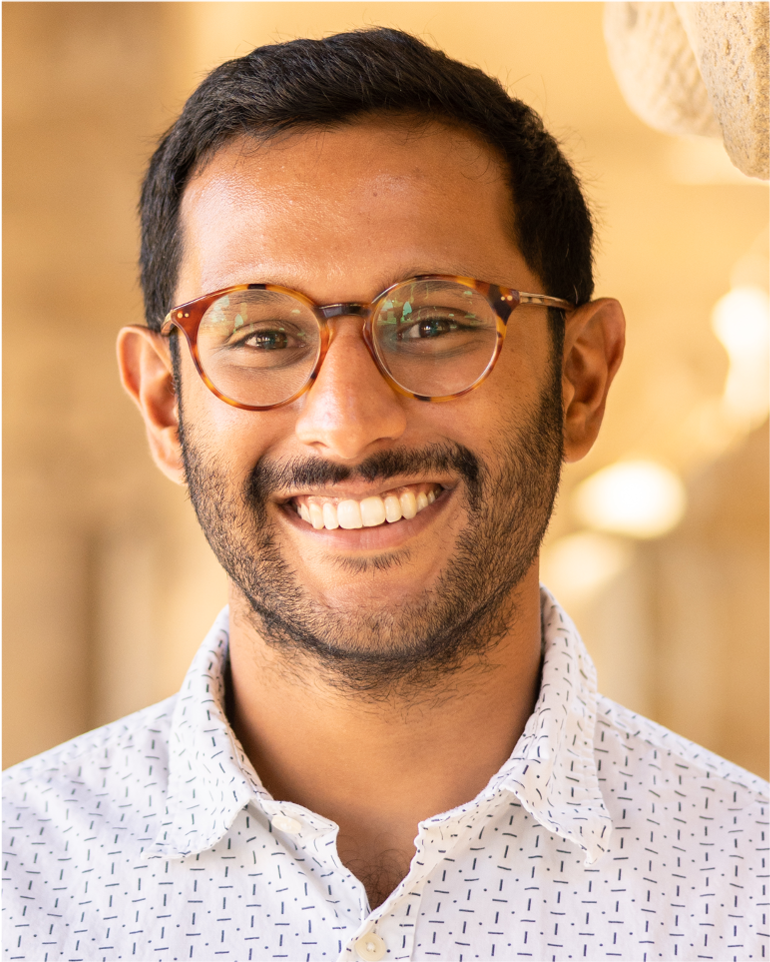} is a Robotics Technologist in the Robotic Surface Mobility Group at  NASA  Jet  Propulsion  Laboratory, California Institute of Technology.
Abhishek received his B.S. in Aerospace Engineering from the University of Michigan - Ann Arbor and his PhD. in Aeronautics and Astronautics from Stanford University.
His research interests lie in leveraging recent advances in nonlinear optimization, machine learning, and control theory towards planning and control for complex robotic systems.
\end{biographywithpic} 

\begin{biographywithpic}
{R. Michael Swan}{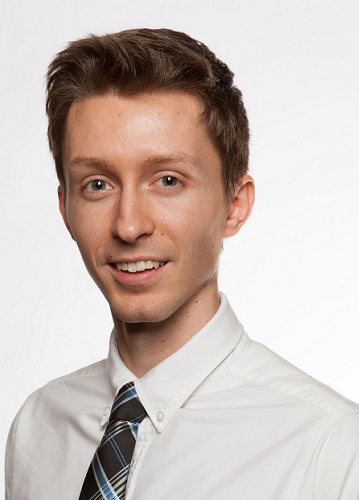} is a Robotics Systems Engineer at  NASA  Jet  Propulsion  Laboratory, California Institute of Technology.
He received his B.S. in Computer Engineering  from  Walla Walla  University  and  his  M.S.  in  Computer  Science from the University of Southern California.
He is interested in robotic surface and aerial autonomy, perception, simulation, and robotic system architecture.
\end{biographywithpic} 

\begin{biographywithpic}
{Hiro Ono}{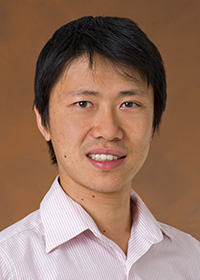} is  the  Group  Leader of the Robotic Surface Mobility Group at  NASA  Jet  Propulsion  Laboratory, California Institute of Technology.
Since he joined JPL in 2013, he has led a number of research projects on Mars rover autonomy, as well as three NIAC studies on Enceladus Vent Explorer and Comet Hitchhiker.
Hiro was a flight software developer  of  M2020’s  Enhanced AutoNav and the lead of M2020 Landing Site Traversability Analysis.
He also led the development of a machine learning-based Martian terrain classifier,  SPOC  (Soil  Property  and  Object  Classification), which won JPL's Software of the Year Award in 2020.
\end{biographywithpic} 

\begin{biographywithpic}
{Shreyansh Daftry}{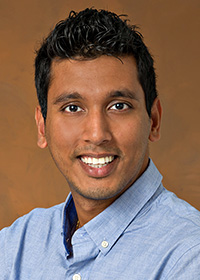} is a Robotics Technologist at NASA Jet Propulsion Laboratory, California Institute of Technology.
He received his M.S. degree in Robotics from the Robotics Institute, Carnegie Mellon University, and his B.S. degree in Electronics and Communication Engineering.
His research interest lies at the intersection of space technology and autonomous robotic systems, with an emphasis on machine learning applications to perception, planning, and decision making.
At JPL, he is the Group Leader of the Perception Systems group, is working on the Mars Sample Recovery Helicopter mission, and has led/contributed to technology development for autonomous navigation of ground, airborne, and subterranean robots.
\end{biographywithpic}

\begin{biographywithpic}
{John Elliott}{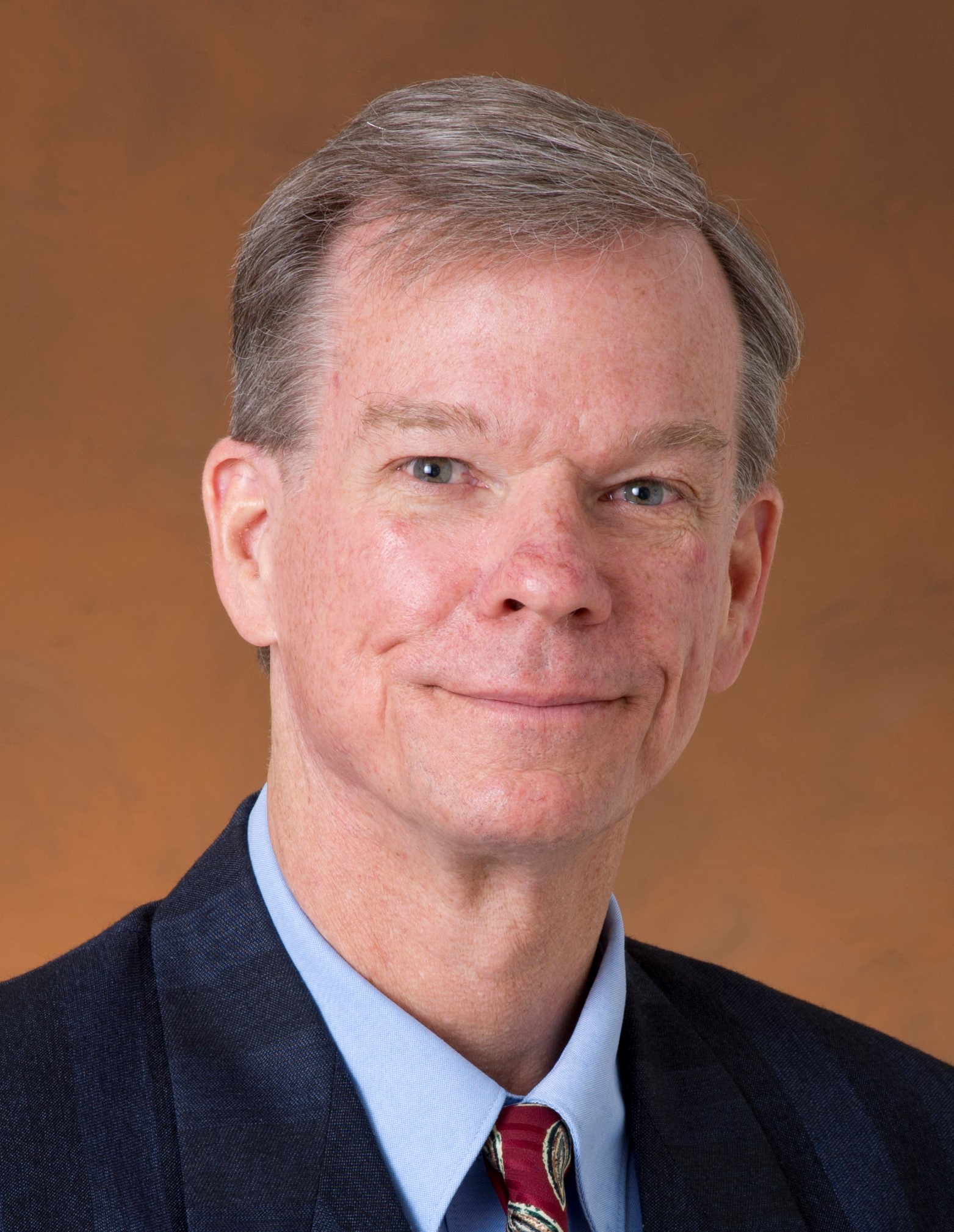} is a principal engineer in JPL’s Mission Concept Systems Development group.  He currently serves as Program Engineer for the Planetary Science Formulation office. His recent tasks have included serving as study lead for the Planetary Decadal Survey’s three lunar rover mission concept studies, Intrepid, INSPIRE, and Endurance, and performing systems engineering and leadership roles on a number of recent Discovery and New Frontiers mission proposals. Mr. Elliott’s past experience includes six years in the terrestrial nuclear power industry with Bechtel Corporation in addition to 30 years in aerospace systems at TRW and JPL.
\end{biographywithpic} 

\begin{biographywithpic}
{Larry Matthies}{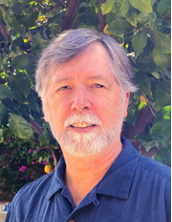} is the technology co-
ordinator in the Mars Exploration Pro-
gram Office at JPL. 
He received B.S., M. Math,and   PhD   degrees   in   Computer   Science   from   the   University   of   Regina (1979),  University  of  Waterloo  (1981), and Carnegie Mellon University (1989).
He  has  been  with  JPL  for  more  than 32 years.
He has conducted technology development  in  perception  systems  for autonomous navigation of robotic vehicles  for  land,  sea,  air,  and  space. 
He supervised  the  JPL  Computer  Vision  group  for  21  years.
He led development of computer vision algorithms for Mars rovers, landers, and helicopters.
He is a Fellow of the IEEE and a member of the editorial boards of Autonomous Robots and the Journal of Field Robotics.
\end{biographywithpic} 

\begin{biographywithpic}
{Deegan Atha}{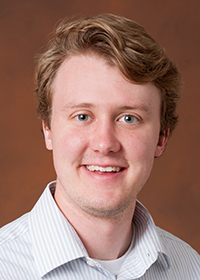}
is a Robotics Technologist within the Perception Systems Group of the Mobility and Robotic Systems Section at the Jet Propulsion Laboratory. He received his B.S. degree from Purdue University in Electrical Engineering and his M.S. in Computer Science from the Georgia Institute of Technology. He is currently the Principal Investigator for the ShadowNav task and leading the semantic perception effort for the DARPA RACER project. His interests are in the infusion of machine learning and robotic perception into autonomous systems operating in unstructured environments.
\end{biographywithpic} 

\end{document}